# Interpretability of Statistical, Machine Learning, and Deep Learning Models for Landslide Susceptibility Mapping in Three Gorges Reservoir Area

Cheng Chen[a], Lei Fan[a],*

[a]Department of Civil Engineering, Design School, Xi'an Jiaotong-Liverpool University, Suzhou, 215000, China

**Abstract**

Landslide susceptibility mapping (LSM) is crucial for identifying high-risk areas and informing prevention strategies. This study investigates the interpretability of statistical, machine learning (ML), and deep learning (DL) models in predicting landslide susceptibility. This is achieved by incorporating various relevant interpretation methods and two types of input factors: a comprehensive set of 19 contributing factors that are statistically relevant to landslides, as well as a dedicated set of 9 triggering factors directly associated with triggering landslides. Given that model performance is a crucial metric in LSM, our investigations into interpretability naturally involve assessing and comparing LSM accuracy across different models considered. In our investigation, the convolutional neural network model achieved the highest accuracy (0.8447 with 19 factors; 0.8048 with 9 factors), while Extreme Gradient Boosting and Support Vector Machine also demonstrated strong predictive capabilities, outperforming conventional statistical models. These findings indicate that DL and sophisticated ML algorithms can effectively capture the complex relationships between input factors and landslide occurrence. However, the interpretability of predictions varied among different models, particularly when using the broader set of 19 contributing factors. Explanation methods like SHAP, LIME, and DeepLIFT also led to variations in interpretation results. Using a comprehensive set of 19 contributing factors improved prediction accuracy but introduced complexities and inconsistency in model interpretations. Focusing on a dedicated set of 9 triggering factors sacrificed some predictive power but enhanced interpretability, as evidenced by more consistent key factors identified across various models and alignment with the findings of field investigation reports. Our comparative analysis highlighted the trade-off between accuracy and interpretability when deciding on the factors to be incorporated, emphasizing the need for balancing accuracy and interpretability in landslide susceptibility modelling.

## 1. Introduction

    Landslides are a common geological hazard that involves the downslope movement of rock or soil under the influence of hydrological, climatic, or human activities factors (Chen and Fan 2023b; Hungr et al. 2014; Saito et al. 2009). Assessing areas prone to landslides is crucial for predicting and mitigating landslide hazards. As a commonly used landslide assessment approach, landslide susceptibility mapping (LSM) evaluates the likelihood of landslide occurrence and delineates landslide-prone areas to form the basis for establishing a landslide management strategy.

    Recent research has extensively investigated the adaptation of data-driven models for LSM, particularly driven by rapid advancements in remote sensing, machine learning (ML) and deep learning (DL) techniques (Ma et al. 2023). Unlike physical-based (Medina et al. 2021) or expert-knowledge-driven (Medina et al. 2021) LSM models, which depend heavily on slope stability analysis or expertise, respectively, data-driven models determine the susceptibility of landslide-prone areas from historical landslide data and various spatiotemporal datasets containing factors that influence landslide occurrence, such as topography, geology, vegetation, land use and climate (Zêzere et al. 2017).

    Data-driven models typically include statistical analysis models (Reichenbach et al. 2018), ML models (Merghadi

---

* Corresponding author.
  *E-mail address:* lei.fan@xjtlu.edu.cn (Lei Fan)

et al. 2020), and DL models (Bui et al. 2020).Commonly used statistical analysis methods include information value (IV) (Chen et al. 2016), weights of evidence (WoE) (Armaş 2012) and frequency ratio (FR) (Li et al. 2017). Compared to statistical analysis methods, ML models, such as support vector machines (SVM) (Huang and Zhao 2018), decision trees (DT) (Guo et al. 2021), logistic regression (LR) (Sun et al. 2021), and artificial neural networks (ANN) (Mehrabi and Moayedi 2021) exhibit improved predictive accuracy of LSM, thanks to their great capabilities in capturing complex nonlinear relationships and establishing more precise links between contributing factors and landslide occurrences. DL models, frequently outperforming conventional ML models, utilize a larger set of parameters to reveal hidden interconnections among multiple contributing factors, thereby effectively modelling complex nonlinear relationships and elevating predictive accuracy (Aslam et al. 2023; Chen and Fan 2023a). However, data-driven models often lack intuitive explanations between input factors and landslide occurrence, posing challenges for users to comprehend the roles of different contributing factors (Russell et al. 2023). Moreover, some sophisticated DL models intensively depend on abundant trainable parameters to accomplish formidable nonlinear fitting capabilities, which may further diminish model interpretability (Li et al. 2022a).

In the context of data-driven LSM models, interpretability aims to explain the contribution of individual input contributing factors to landslide occurrences (Molnar 2020). In general, interpretability can be categorized into global and local interpretations (Lundberg et al. 2020). Global interpretation analyses how the model responds to different input factors across its full range of input factors values and how the input factors impact the prediction of the landslide susceptibility, providing insights into the overall feature importance, global model behaviours, etc. However, global interpretation can only identify the overall critical factors across the whole study area, unable to determine the important factors for specific landslides. Thus, in most situations, global interpretation is deemed insufficient for explaining potential local landslides. In contrast, local interpretation elucidates the predictive behaviours by analysing the influence of site-specific input factors, which exhibit substantial differences across various spatial locations, on the model's predictions for individual instances or localized regions. More precisely, it distinctly aims to clarify the model's responses to these specific inputs within a localized spatial or geological context. For example, Zhou et al. (2022) employed the SHapley Additive exPlanations (SHAP) technique as a local interpretability approach, in conjunction with eXtreme Gradient Boosting (XGBoost) as the prediction model, to precisely analyse and evaluate the factors influencing the occurrence of the Wushanping landslide. This method facilitated a more intricate comprehension of the precise contribution of individual input factors to the prediction of landslide susceptibility.

While data-driven LSM models have consistently improved prediction accuracies by implementing sophisticated algorithms (Lv et al. 2022), their interpretability is still limited. Despite attempts in some studies to address this limitation by employing global interpretation methods for identifying influential factors (Lu et al. 2023; Lv et al. 2023; Sun et al. 2023; Zhou et al. 2022), and more recently, an increasing use of local interpretation methods in specific LSM studies to inspect factors' contribution at a local level (Al-Najjar et al. 2023; Fang et al. 2023; Inan and Rahman 2023; Pradhan et al. 2023; Zhang et al. 2023), LSM model interpretability remains an underexplored area. Additionally, there is a notable absence of systematic evaluations and comparisons of interpretability across various data-driven LSM models and interpretation mechanisms.

To address the aforementioned research gaps, our study is dedicated to exploring and comparing interpretability of various interpretation methods when incorporated into data-driven LSM models. This enables us to gain a better understanding of the explanatory consistency between different LSM models coupled with relevant interpretation methods for global interpretation, and the explanatory accuracy for local specific landslide causes informed by field investigation reports. Specifically, we investigated three types of data-driven LSM models: IV, WoE, and FR as statistical models, LR, SVM, and XGBoost as ML models with SHAP as the interpretation method, and Long Short-Term Memory (LSTM) and Convolutional Neural Network (CNN) as DL models with SHAP and Deep Learning Important Feature attribution (DeepLIFT) as the interpretation methods. For local interpretation, we employed the following methods: Local Interpretable Model-Agnostic Explanations (LIME), SHAP and DeepLIFT. This enables evaluation of prediction behaviour and interpretation at individual landslide locations.

## 2. Study Area and Data

*2.1. Study Area*

The study area pertains to the Three Gorges Reservoir Area (TGRA) in China, consisting of latitudinal coordinates spanning from 28°52' to 31°74' N and longitudinal coordinates ranging from 105°72' to 111°68' E, as shown in Fig.1. TGRA straddles two distinct natural geographical units, bisected by Bai Di Cheng in Fengjie County. The eastern section is characterized by tall mountains and deep valleys, while the western portion is situated within the Sichuan Basin, featuring folded hills formed at the confluence with the western Hubei Mountains. The primary geological composition comprises sedimentary strata, with minor occurrences of magmatic and metamorphic basement rocks. The climate is classified as subtropical monsoonal, marked by notable seasonal variations in both temperature and precipitation (Li et al. 2022b).

The unique topography, rainfall patterns, and geology of the Three Gorges region make it an ideal place to study geologic hazards and their distribution. TGRA's complex topography, coupled with heavy seasonal rainfall and diverse geology, inherently makes it susceptible to landslide occurrences. Moreover, the construction of the Three Gorges reservoir has significantly altered slopes characteristics and hydrological patterns, exacerbating landslide risks. Potential landslides pose a direct threat to the communities and infrastructure within TGRA. Therefore, it is important to accurately locate potential landslide hazards for actions to effectively mitigate risks.

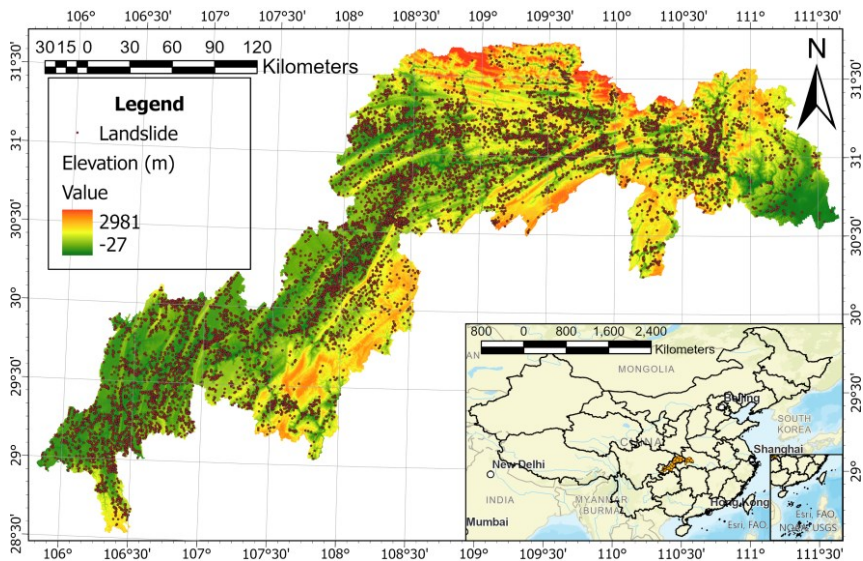

Fig. 1. Location of TGRA and its landslide inventory map, with the elevation data from the ALOS satellite and the base map from ESRI.

*2.2. Landslide inventory*

The landslide inventory (shown in Fig.1) was compiled from diverse data sources including historical records and satellite image analysis, which allowed accurate mapping of 10,819 landslide events. To complement the dataset for LSM, non-landslide locations were systematically generated to serve as negative samples. Selecting negative samples from ostensibly stable terrain is a conventional approach in landslide susceptibility studies (Zhou et al. 2021). The initial number of non-landslide points generated is four times the number of landslide points in the study area, determined through trial and error. For these initial non-landslide samples, a minimum distance of 3 kilometres from their nearest landslide points is applied to ensure spatial independence. Ultimately, this process produced a total of 11,000 well-distributed non-landslide samples used in this study. Balancing the landslide and non-landslide samples helps prevent overfitting in data-driven models by providing equal representation of hazardous and stable terrain.

## 2.3. Contributing and Triggering Factors for the TGRA

The selection of appropriate contributing factors is crucial for ensuring the accuracy of LSM. After reviewing landslide-related factors documented in prior LSM studies (Sun et al. 2022; Xiao et al. 2023; Xiao et al. 2019; Zhang et al. 2017) and considering data availability, we have identified 19 commonly used contributing factors for LSM modelling in our study. These factors (denoted by C) are summarised in Table 1 and include Elevation, Slope, Aspect, Plan Curvature, Profile Curvature, Surface cut depth, terrain roughness index (TRI), Landform, Lithology, Distance to fault, Land use, Distance to road, normalized difference vegetation index (NDVI), stream power index (SPI), sediment transport index (STI), topographic wetness index (TWI), Distance to stream, Peak rainfall intensity, and Average rainfall intensity. The maps depicting these 19 contribution factors for the TGRA area are illustrated in Figure 2, with the corresponding source data used to create these maps provided in Table 1. The inclusion of these contributing factors stems from the likely statistical relevance of spatial-temporal data to landslides, often under the assumption that incorporating more data types could enhance the predictive accuracy of models across test datasets. However, a main issue with this approach arises from the fact that some of these contributing factors may not be directly relevant to the mechanisms underlying landslide failures. Consequently, this could lead to a misinterpretation of the roles of various factors in triggering landslides.

As this study investigates the interpretability of various LSM models, it deliberately chooses specific triggering factors (referred to as T in Table 1) from the contributing factors, which are considered to exert a direct influence on landslides. These triggering factors, a subset of contributing factors, are identified based on-site survey reports of landslides (Chen et al. 2021; Huang et al. 2018; Li et al. 2019; Tang et al. 2019; Yao et al. 2019) and from a geotechnical engineering perspective. They include Slope, Lithology, Distance to fault, Distance to road, NDVI, TWI, Distance to stream, Peak rainfall intensity, and Average rainfall intensity. Slope and lithology determine the inherent stability conditions of soil strata. Distance to fault, distance to road, and NDVI reflect the influence of geological discontinuities, anthropogenic interventions, and vegetation cover on the mechanical properties and stability of slopes. TWI and distance to stream represent hydrological factors that may elevate pore water pressures and induce landslides by reducing the effective stress within the soil mass. Peak rainfall intensity and average rainfall intensity signify dynamic climatic factors that can trigger landslides through infiltration and saturation of slope materials.

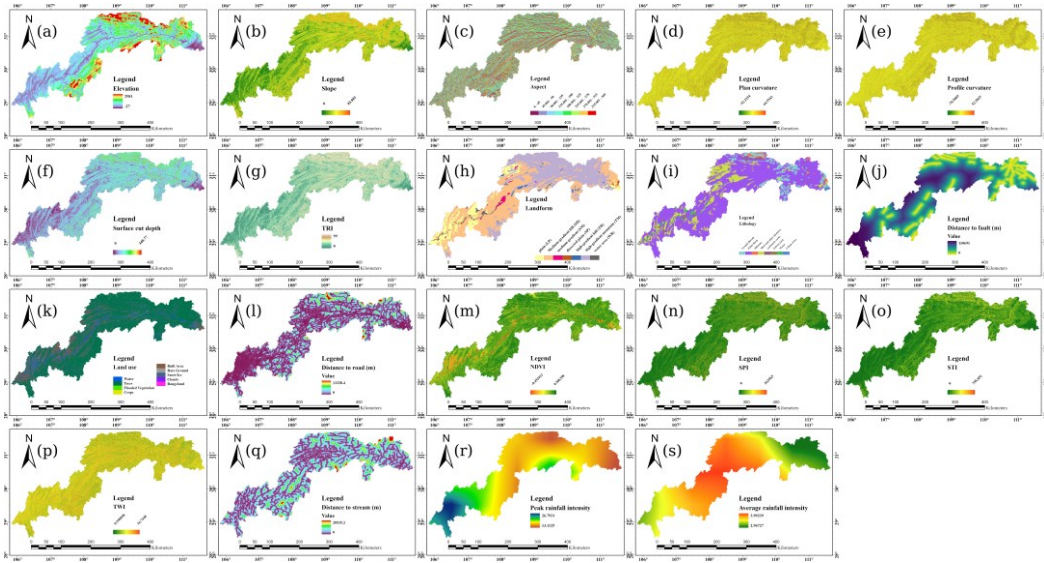

Fig. 2. Maps depicting contributing factors for LSM: (a) Elevation, (b) Slope, (c) Aspect, (d) Plan Curvature, (e) Profile Curvature, (f) surface cut depth, (g) TRI, (h) Landform, (i) Lithology, (j) Distance to fault, (k) Land use, (l) Distance to road, (m) NDVI, (n) SPI, (o) STI, (p) TWI, (q) Distance to stream, (r) Peak rainfall intensity, (s) Average rainfall intensity.

Table 1: Factors related to landslide occurrence identified from previous LSM studies and field surveys.

| Landscape Characteristics | Factor | Mechanism Type | Data source |
|---|---|---|---|
| Topographic | Elevation | C | OpenStreetMap (https://download.geofabrik.de/) |
| | Slope | C, T | Derive from DEM |
| | Aspect | C | Derive from DEM |
| | Plan Curvature | C | Derive from DEM |
| | Profile Curvature | C | Derive from DEM |
| | Surface cut depth | C | Derive from DEM |
| | TRI | C | Derive from DEM |
| Geological Composition | Lithology | C, T | China Geological Survey Geological Cloud (https://geocloud.cgs.gov.cn) |
| | Distance to fault | C, T | China Geological Survey Geological Cloud (https://geocloud.cgs.gov.cn) |
| | TWI | C, T | Derive from DEM |
| Geomorphology | Landform | C | China Geological Survey Geological Cloud (https://geocloud.cgs.gov.cn) |
| Hydraulic | SPI | C | Derive from DEM |
| | STI | C | Derive from DEM |
| | Distance to stream | C, T | OpenStreetMap (https://download.geofabrik.de/) |
| Human activity | Land Use | C | Landsat 8 Product from USGS (https://earthexplorer.usgs.gov/) |
| | NDVI | C, T | Landsat 8 Product from USGS (https://earthexplorer.usgs.gov/) |
| | Distance to road | C, T | OpenStreetMap (https://download.geofabrik.de/) |
| Climatic Conditions | Peak rainfall intensity | C, T | China Meteorological Bureau (http://www.cma.gov.cn/) and Google earth engine |
| | Average rainfall intensity | C, T | China Meteorological Bureau (http://www.cma.gov.cn/) and Google earth engine |

Conditioning Factor (C): Referring to any factors that statistically relate to landslides.
Triggering Factor (T): Referring to specific factors that directly cause or drive landslides.

Fig. 3 illustrates the statistical distributions of both landslide and non-landslide samples. Notably, elevation, average rainfall intensity and slope exhibit greater degree of variation for landslide samples compared to non-landslide samples. Additionally, landslide samples tend to occur at lower elevations, closer to rivers, and with shorter distances from roads. Furthermore, distinctive variations are evident in landslide samples concerning cutting depth, distance from fault, TRI and TWI, distance to stream and distance to road.

Developing a landslide susceptibility model requires attention to the issue of multicollinearity. Multicollinearity refers to the strong correlation among landslide contributing factors, leading to excessive bias and unreliable model outcomes. Table 2 presents the Variance Inflation Factors (VIF) for each factor, serving as a metric to weigh the extent of multicollinearity. Generally, VIF values below 10 are considered indicative of the absence of multicollinearity (Gu et al. 2023). In our case, all values in Table 2 fall below this threshold.

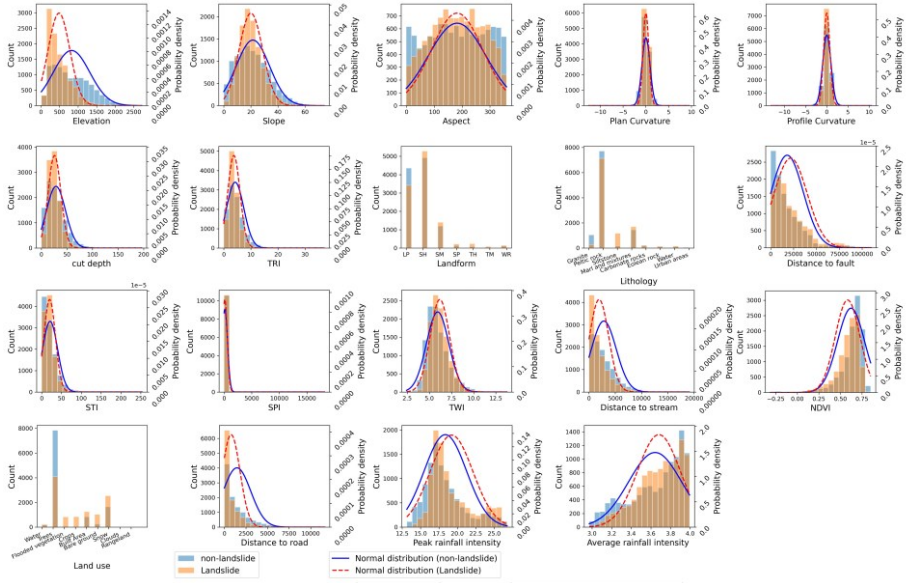

Fig. 3. Distribution of factor values at landslide and non-landslide locations.

Table 2. VIF values for each contributing factor.

| Factor | VIF |
|---|---|
| Elevation | 7.0494 |
| Slope | 1.0751 |
| Aspect | 4.0587 |
| Plan Curvature | 2.3141 |
| Profile Curvature | 1.3434 |
| Cut Depth | 3.2063 |
| TRI | 6.6068 |
| Landform | 8.9770 |
| Lithology | 6.5506 |
| Distance to fault | 5.7977 |
| STI | 5.6325 |
| SPI | 2.6556 |
| TWI | 5.2835 |
| Distance to stream | 5.5343 |
| NDVI | 2.3436 |
| Land use | 5.3439 |
| Distance to road | 2.9109 |
| Peak rainfall Intensity | 4.0635 |
| Average rainfall Intensity | 6.2020 |

## 3. Methodology

Fig. 4 illustrates the methodological flowchart employed in this study, outlining the workflow comprising data preparation, model construction, and model interpretation stages. In the data preparation stage, a landslide inventory is compiled, along with a set of 19 contributing factors and 9 triggering factors, which have been elaborated in Section 2. The model construction stage involves developing and validating LSM models using statistical (IV, WoE, FR), machine learning (LR, SVM, XGBoost), and deep learning (LSTM, CNN) techniques. The generated LSM maps from these diverse models are compared to assess their performance. The model interpretation stage focuses on both global and local interpretability. As depicted in Fig. 4, various interpretation methods are paired with the employed LSM models to obtain global factor importance and local factor importance. The factor importance derived from different combinations of LSM models and interpretation methods is utilized to understand the contributions of various factors to landslide susceptibility predictions and to compare the consistency of their interpretability across different combinations. Section 3.1 provides a detailed description of the specific models employed in this study, while Section 3.2 elaborates on model interpretation methods.

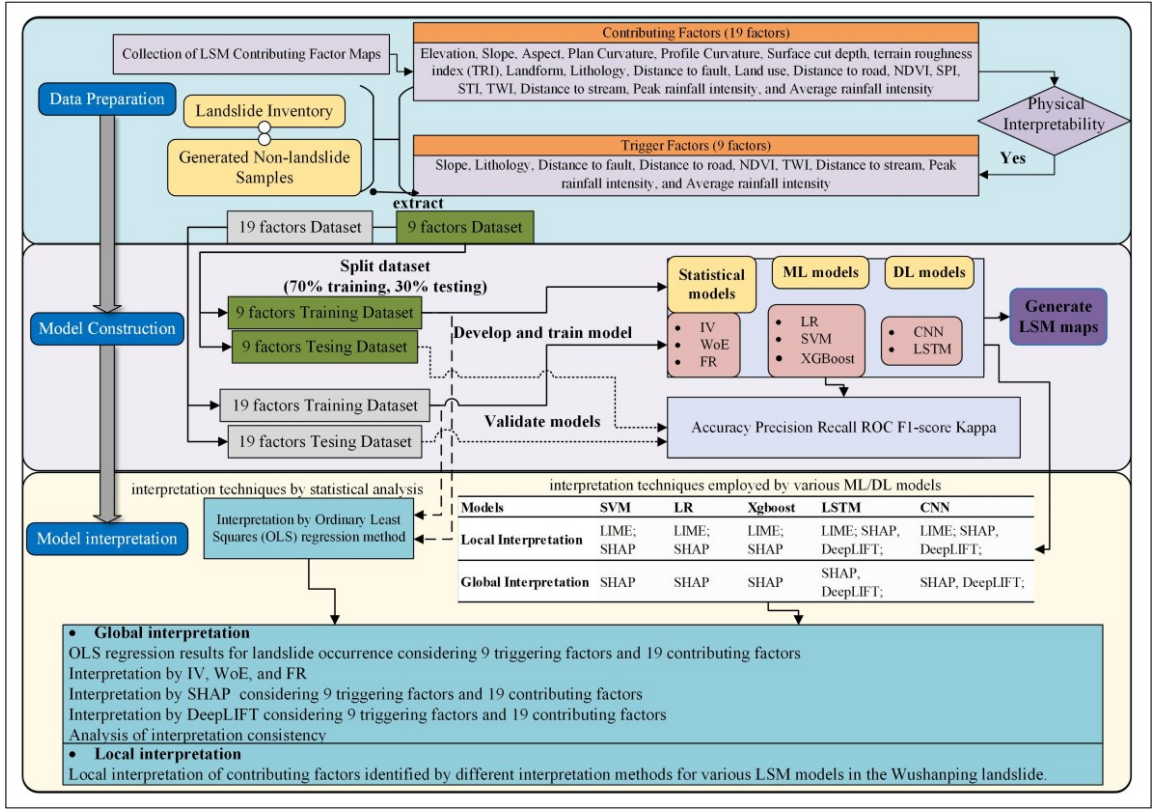

Fig. 4. Methodological flowchart for investigating interpretability across LSM models and interpretation methods

## 3.1. Model

### 3.1.1. Statistical analysis

- **IV**

The IV method, grounded in information theory, was originally proposed by Yin (1988) and further refined by Westen (1993). In this method, the IV for each parameter class is calculated as the natural logarithm of the landslide density within a class divided by the landslide density across the entire map (shown in Eq. (1)). The weight allocated to a parameter class is determined by this IV value. A higher IV value corresponds to a higher weight, signifying a greater possibility of a landslide occurring within a specified area (Wang et al. 2019). The IV method ascertains their significance of the contributing factors in determining landslide susceptibility.

$$I_i = \ln\left(\frac{Densclass}{Densmap}\right) = \ln\left(\frac{N_{pix}(S_i)}{N_{pix}(N_i)} \bigg/ \frac{\sum_{i=1}^{n} N_{pix}(S_i)}{\sum_{i=1}^{n} N_{pix}(N_i)}\right) \quad (1)$$

Where $I_i$ denotes the weight assigned to a factor class; *Densclass* represents the landslide density within a factor class; *Densmap* is the total landslide density across the entire map; $N_{pix}(S_i)$ is the number of landslides in a specific factor class; $N_{pix}(N_i)$ is the number of pixels in a given factor class; and $n$ is the number of classes in a factor map.

Therefore, the landslide susceptibility index (LSI) for each pixel was computed by summing the information values of each factor class as Eq. (2):

$$LSI = \sum_{j=1}^{n} I_i \quad (2)$$

- **WoE**

The WoE method is a statistical method that uses the Bayesian probability model. It was originally developed for mineral potential assessment (Bonham-Carter et al. 1988), and WoE method involves assigning weights to landslide contributing factors based on their spatial relationship with landslides, thus facilitating a comprehensive landslide susceptibility assessment. The essence of WoE method lies in deducing positive ($W+$) and negative ($W-$) weights for each contributing factor based on the presence or absence of landslides ($A$) or ($\overline{A}$) in a specific area, represented by the Eq. (3) and Eq. (4).

$$W^+ = \ln \frac{P\{B|A\}}{P\{B|\overline{A}\}} \quad (3)$$

$$W^- = \ln \frac{P\{\overline{B}|A\}}{P\{\overline{B}|\overline{A}\}} \quad (4)$$

Here, $P$ signifies the probability ratio, and ln denotes the natural logarithm. $B$ represents the categories or classes of the factor, and $\overline{B}$ represents the non-members or other classes. $W^+$ and $W^-$ reveal the positive and negative correlations, respectively, between landslides and a factor (Bourenane et al. 2016). The weight contrast $C$, deemed as a measure reflecting the overall importance of a factor map class, is calculated as the difference between these weights $C = W^+ - W^-$. The standard deviation of is computed as Eq. (5).

$$S(C) = \sqrt{S^2(W^+) + S^2(W^-)} \quad (5)$$

Where $S^2(W^+)$ and $S^2(W^-)$ denote the variance of the $W^+$ and $W^-$ weights, respectively.

The studentized contrast $W_f$, serving as a confidence measure, is defined as Eq.(6) (Kayastha et al. 2012), and LSI is thus determined by the Eq.(7).

$$W_f = \frac{C}{S(C)} \quad (6)$$

$$LSI = \sum_{j=1}^{n} W_f \quad (7)$$

- **FR**

The FR method, as presented by Choi et al. (2012), leverages the spatial relationship between landslides and contributing factors. The FR simplifies the calculation of probabilities that include both dependent and independent variables, using categorized maps. The FR model is well-regarded for its simplicity and straightforward interpretation of results (Yilmaz 2009). By quantifying associations between regions of historical landslide incidence and their contributing factors, the FR model facilitates effective delineation of landslide susceptibility across candidate areas. The FR is calculated as the ratio of landslides within a specific class to the total landslides, normalized by the ratio of the entire class to the overall area (Nourani et al. 2013). The $FR_{ij}$ is given by Eq. (8).

$$FR_{ij} = \frac{FL_{ij}}{FN_{ij}} \quad (8)$$

where $FR_{ij}$ signifies the FR in class $i$ for factor $j$, $FL_{ij}$ represents the ratio of landslide-affected pixels in class $i$ for factor $j$ to the total landslides in all classes for factor $j$, and $FN_{ij}$ is the ratio of the total pixels in class $i$ to all pixels in the map.

A higher $FR_{ij}$ indicates a stronger correlation between landslide occurrence and the factor class, while a lower ratio implies a weaker association between factor class and landslide occurrence (Lee and Pradhan, 2006; Yalcin et al., 2011). To compute the LSI, Eq. (9) is employed.

$$LSI = \sum_{j=1}^{n} FR_{ij} \tag{9}$$

### 3.1.2. ML models

- **LR**

LR is a widely used ML model that establishes a quantitative relationship between factors and landslide occurrence through regression analysis. It fits data with a curve, often a sigmoid function, aiming to minimize the distance between the curve and the data. The logistic regression model (Eq. (12)) is derived from the logistic function (Eq. (10)), presenting a weighted linear model (Eq. (11)) with correlation coefficients ($\beta_i$) and an intercept ($\beta_0$).

$$f(z) = \frac{1}{1+e^z} \tag{10}$$

$$h_w(x) = \beta_0 + \sum_{i=1}^{n} \beta_i x_i \tag{11}$$

$$f_\omega(x) = \frac{1}{1+e^{h\omega(x)}} \tag{12}$$

- **SVM**

SVM is also a popular ML model in LSM, excelling in generalization ability, particularly with small sample problems. It addresses local optimum and dimensionality problems by mapping data from the input space to a high-dimensional feature space, followed by nonlinear regression analysis to determine the optimal classification hyperplane. The objective function of SVM aims to minimize the margin distance between sample points and the hyperplane (Eqs. (13–14)).

$$\text{Minimize } \frac{1}{2}\|w\|^2 + C \sum_{i=1}^{l} \xi_i + \xi_i^* \tag{13}$$

$$\text{Subject to: } -\varepsilon - \xi_i \leq \omega^T \Phi(x^{(i)}) + b - y^{(i)} \leq \varepsilon + \xi_i^*, \forall i \tag{14}$$

The radial basis function (RBF) (Eq. (15)) is employed as the kernel function in this study.

$$\Phi(X_i, X) = \exp\left(-\frac{\|X-X_i\|^2}{2\sigma^2}\right) \tag{15}$$

- **XGBoost**

XGBoost is an advanced boosting framework based on decision tree algorithms. It is designed for efficient and distributed training, making it exceptionally fast and scalable. XGBoost employs a gradient boosting learning approach, where it discretizes continuous features into bins and utilizes them to construct trees.

The objective function in XGBoost aims to minimize a specific loss function, varying based on the task (e.g., regression, classification). For instance, in binary classification (i.e., landslide classification), the binary logistic regression loss function is commonly used. The overall objective function can be written as Eq. (17). The objective function aims to minimize a loss function $L(\theta)$ which involves predictions deviation $L(y\_i, f(x\_i;\theta))$ and regularization $\Omega(f_k)$. The regularization term $\Omega(f_k)$ encourages model simplicity and is defined as Eq. (16), where $T$ is the number of leaves and $w_j$ is the leaf weight. The overall objective is to find optimal parameters $\theta$ by iteratively improving predictions using weak learners (trees).

$$\Omega(f_k) = \gamma T + \frac{1}{2}\lambda \sum_{j=1}^{T} w_j^2 \tag{16}$$

$$L(\theta) = \sum_i \mathcal{L}(y_i, f(x_i;\theta)) + \sum_{k=1}^{K} \Omega(f_k) \tag{17}$$

### 3.1.3. DL models

- **CNN**

A standard CNN architecture comprises fundamental elements: a convolutional layer, a pooling layer, and a fully connected layer (Chen and Fan 2023b). CNN serves as an encoder, proficient in capturing significant features without the need for intricate rules, making it a formidable tool for discriminative tasks. The primary objective of the convolutional layer is to extract distinctive features from the input factor. By integrating multiple convolutional layers, the CNN iteratively learns increasingly sophisticated representations, starting from low-level inputs. The 1D CNN adopted in this study, which employs a unidirectional approach to shift the convolutional kernel. MaxPooling, a prevalent downsampling technique, is employed to reduce the dimensions of the feature map while preserving its depth. The activation function in the convolutional layer, the linear exponential unit, accelerates convergence and enhances the model's resilience.

Compared to traditional statistical or machine learning models in LSM, CNNs autonomously extract relevant features from input factor, reducing the need for manual feature engineering. Additionally, CNNs can handle large and complex datasets effectively, making them highly suitable for LSM tasks.

- **LSTM**

LSTM, a specialized variant of Recurrent Neural Networks (RNNs), effectively addresses the vanishing gradient problem, expediting convergence. The fundamental LSTM structure encompasses three crucial components: the input gate, forget gate, and output gate, which collectively regulate information flow in and out of the memory cell.

LSTM excels at capturing long-term dependencies in sequential data, making it particularly effective for time-series prediction and analysis. Compared to traditional statistical models in LSM, LSTM is specialized in handling sequential data, effectively capturing temporal dependencies and long-term patterns. These advancements empower LSM by enhancing predictive accuracy and providing a deeper understanding of landslide susceptibility in complex geological settings.

### 3.1.4. Model Hyperparameters Configuration

Table. 3 outlines the hyperparameter configurations utilized in our study for models of SVM, LR, XGBoost, CNN, and LSTM. It succinctly summarizes the parameter ranges explored and highlights the best parameters identified through systematic experimentation and optimization. Employing the Tree-structured Parzen Estimator hyperparameter optimization, these configurations achieve effective fine-tuning, thereby optimizing predictive performance of models used within this study.

Table 3. Summary of the model parameters used.

| Model | Parameters range | Best Parameters |
|---|---|---|
| SVM | C = (0.1, 1, 5, 10, 50, 100, 180) | C = 5 |
| | kernel = ('linear', 'poly', 'rbf') | Kernel = 'rbf' |
| LR | C = (0.1, 1, 5, 10, 50, 100, 180) | C = 50 |
| | penalty = 'l1' or 'l2' | Penalty = 'l1' |
| | solver = 'lbfgs', 'liblinear' | Solver = 'lib-linear' |
| XGboost | n_estimators = [50,100, 150, 200] | n_estimators = 150 |
| | learning rate = [0.01,0.1] | gamma = 0.08 |
| | max_depth = [2,5,10,15,20] | max_depth = 15 |
| CNN | filters = [16, 32, 64, 128] | Filters = 64 |
| | kernel_size = (3, 3), (5,5), (7,7) | kernel_size = (3, 3) |
| | dropout = [0.1, 0.2, 0.3] | Dropout = 0.2 |
| | activation = 'relu', 'tanh' | Activation = 'relu' |
| LSTM | units = [50, 100, 200, 300] | Units = 100 |
| | dropout = [0.1, 0.2, 0.3] | Dropout = 0.2 |
| | activation = 'relu', 'tanh' | Activation = 'tanh' |

## 3.2. Interpretation Methods for Data-Driven Models

In LSM, comprehending the importance and influence of contributing factors is important. Numerous interpretation techniques can be used to uncover the underlying mechanism behind model predictions. Such techniques can be categorized as local or global interpretation methods. The differences between local and global interpretation methods are depicted in Table 4. Specifically, local interpretation focuses on explaining individual model predictions, aiming to unravel the reasoning behind specific outputs by examining isolated input factors in detail. Approaches such as LIME, SHAP and DeepLIFT can provide local model interpretation. In contrast, the aim of global interpretation is to comprehensively understand the overall behavior and decision-making process of the model across the entire dataset.

Table 4. Difference between local and global interpretation methods.

| Interpretation | Local Explanation | Global Explanation |
| --- | --- | --- |
| Scope | Explain the model's prediction for a single instance or a small subset of instances. | Explain the model's overall behavior across the entire dataset. |
| Goal | Understand how each input factor contributes to the model's prediction for a specific instance. | Identify the general importance of input factors across the entire model. |
| Explanation | Provide the contribution or importance of each input factor for a specific model prediction. | Provide an overview of the factor importance and interactions within the model across the entire dataset. |
| Methodology | LIME; SHAP; DeepLIFT. | Statistical Analysis; SHAP; DeepLIFT. |

- **Statistical Analysis**

Statistical analysis facilitates comprehensive global interpretation of landslide susceptibility models from multiple perspectives. By examining distributions of input non-landslide and landslide samples, these techniques collectively assess factor importance, identify key influences and their impacts, and reveal underlying mechanisms driving landslides. By synthesizing these statistical tools, interpretive insights are obtained regarding the overall model from quantitative interrelationships, which illuminate the most influential factor effects for an enhanced comprehension of the contributing factors implicated in landslide hazard prediction.

- **LIME**

LIME explains predictions of ML model at a local level by approximating the model's behavior around the prediction using a simple linear model. It generates an explanation that is a linear weighted combination of the features to describe the contribution of each feature to the prediction. In the LIME methodology, given $f$ as the predict model, and $g$ as the interpretable model (e.g., linear regression), the model $f$ is approximated locally around a particular instance $x$ by optimizing the interpretable $g$ model through minimizing a loss function $L(f, g, \pi_x)$ as formulated in Eq. (18).

$$L(f, g, \pi_x) = \sum_z \pi_x(z)[f(z) - g(z)]^2 \tag{18}$$

where $\pi_x(z)$ is the proximity kernel weighing the instances $z$ based on their similarity to $x$.

- **SHAP**

SHAP values offer an equitable distribution of each feature's contribution to the prediction, ensuring that the total contributions accurately account for the difference between the actual and expected prediction. SHAP values are computed by calculating the contribution of each feature in every possible combination, considering all feasible subsets. For a prediction $f(x)$, the SHAP value for feature $i$ is defined as Eq. (19).

$$\phi_i(f, x) = \sum_{S \subseteq N \setminus \{i\}} \frac{|S|!(|N|-|S|-1)!}{|N|!} [f(S \cup \{i\}) - f(S)] \tag{19}$$

where $N$ is the set of all features, $S$ is a subset of $N$ without $i$, and $f(S)$ is the model output for subset $S$.

- **DeepLIFT**

DeepLIFT is a DL model interpretation method that computes attribution scores reflecting the importance of each input feature based on propagation rules. Specifically, DeepLIFT assigns contributions based on the differences in activation from the reference. For a neuron $j$ in layer $l$, the contribution $C_j^l$ is calculated as:

$$C_j^l = \left(a_j^l - r_j^l\right) \times \Sigma \left(\frac{\partial a_j^l}{\partial a_i^{l-1}}\right) \tag{20}$$

where $a_j^l$ is the activation of neuron $j$ in layer $l$, $r_j^l$ is the reference activation, and the summation runs over all neurons $i$ in the layer $l-1$.

### 3.3. Evaluation matrix

This study employs multiple quantitative metrics (Eq. (22-27)) to comprehensively evaluate the performance of the model in predicting landslide susceptibility, including the Area Under the Curve (AUC), accuracy, precision, recall, F1-score, and Kappa. These metrics are calculated based on the confusion matrix, which consists of true positives (TP), true negatives (TN), false positives (FP), and false negatives (FN). TP represents the number of correctly predicted landslide instances, TN represents the number of correctly predicted non-landslide instances, FP represents the number of non-landslide instances incorrectly predicted as landslides, and FN represents the number of landslide instances incorrectly predicted as non-landslides. P and N in Eq. (22) represent the total number of positive (landslide) and negative (non-landslide) instances, respectively.

AUC is an important indicator for measuring the overall accuracy of the model, with a value ranging from 0.5 to 1. A higher value indicates a better performance of the model. Accuracy reflects the model's ability to correctly classify data, with a higher value indicating a lower chance of misclassification. Precision signifies the model's ability to minimize false positives. A higher precision value implies a lower likelihood of falsely identifying non-landslide locations as landslides. Recall measures the model's effectiveness in identifying true landslide instances, with a higher value indicating a better ability to capture real landslide situations and minimize false negatives. F1-score provides a balanced evaluation of precision and recall, with a higher value indicating a better balance. Lastly, Kappa measures the level of agreement between the model's classifications and random chance, with a higher value indicating a better performance of the model. It is calculated using Eq. (27), where $P_{Acc}$ is the observed agreement between the model's classifications and the actual landslide occurrences, and $P_e$ is the random correct rate (Chen et al. 2018). By considering these metrics comprehensively, we can more accurately evaluate the model's performance in predicting landslide susceptibility.

$$AUC = \frac{(\Sigma TP + \Sigma TN)}{(P+N)} \tag{22}$$

$$Accuracy = \frac{TP+TN}{TP+FP+TN+FN} \tag{23}$$

$$Recall = \frac{TP}{TP+FN} \tag{24}$$

$$Precision = \frac{TN}{TN+FP} \tag{25}$$

$$F1_{score} = 2 \times \frac{Precision \times Recall}{Precision + Recall} \tag{26}$$

$$Kappa = \frac{P_{Acc} - P_e}{1 - P_e} \tag{27}$$

## 4. Results

### 4.1. Model performance

The performances of various ML and DL models are summarized in Table 5. Two sets of performance results are presented, corresponding to models trained using 19 contributing factors and 9 triggering factors, respectively. When using 19 contributing factors, CNN achieved the highest accuracy of 0.8447, followed by XGBoost (0.8420) and SVM (0.8364). The performance of DL and ML models varied across different evaluation metrics. CNN achieved the highest precision and AUC among all models, while SVM slightly surpassed others in recall. CNN, XGBoost, and SVM obtained comparable results in terms of F1 score. In the machine learning category, XGBoost outperformed LR in all metrics, leading in precision, F1 score, and AUC. In summary, CNN and XGBoost exhibited the best overall performance across various metrics, highlighting the advantages of DL and ensemble ML learning for this task.

As indicated in Table 5, when using 9 triggering factors, the overall model performance decreased compared to using 19 contributing factors. XGBoost achieved the highest accuracy of 0.8177 amongst all considered models, while CNN achieved an accuracy of 0.8048, outperforming the other DL method namely LSTM. The precision, recall, F1 score, and AUC values also exhibited a similar trend, with XGBoost and CNN generally outperforming other models.

In both data sets (i.e., 19 factors and 9 factors), CNN, XGBoost, LSTM and SVM showed strong model performance, demonstrating their potential for effectively classifying the landslide dataset. However, the reduced performance when using 9 triggering factors suggests that the additional contributing factors captured by the 19-factor set provide valuable supplementary information for improving landslide susceptibility prediction.

Table 5. Values of the performance metrics for the ML and DL models considered, using 19 factors and 9 factors.

| Models | Number of factors | Accuracy | Precision | Recall | F1 score | AUC | Kappa |
|---|---|---|---|---|---|---|---|
| SVM | 19 | 0.8364 | 0.8353 | **0.8357** | **0.8355** | 0.8364 | 0.6728 |
|  | 9 | 0.7837 | 0.7704 | 0.8045 | 0.7871 | 0.7839 | 0.5676 |
| LR | 19 | 0.7031 | 0.6828 | 0.7519 | 0.7157 | 0.7034 | 0.4066 |
|  | 9 | 0.6487 | 0.7047 | 0.6313 | 0.6660 | 0.6490 | 0.2979 |
| XGboost | 19 | 0.8420 | 0.8608 | 0.8136 | **0.8365** | 0.8418 | 0.6839 |
|  | 9 | 0.8177 | 0.8379 | 0.7878 | 0.8121 | 0.8175 | 0.6353 |
| CNN | 19 | **0.8447** | **0.8711** | 0.8069 | **0.8378** | **0.8445** | **0.6893** |
|  | 9 | 0.8048 | 0.8085 | 0.7956 | 0.8020 | 0.8047 | 0.6095 |
| LSTM | 19 | 0.8314 | 0.8406 | 0.8239 | 0.8321 | 0.8321 | 0.6630 |
|  | 9 | 0.7739 | 0.7710 | 0.7752 | 0.7731 | 0.7739 | 0.5477 |

### 4.2. Comparison of susceptibility maps by different LSM models

Fig. 5 and Fig. 6 illustrates landslide susceptibility maps generated using different LSM models, including IV, FR, WoE, LR, SVM, XGBoost, CNN, and LSTM. For each model, two sets of maps are displayed, corresponding to the results obtained using 19 contributing factors and 9 triggering factors, respectively. Employing the (Ke et al. 2023), these maps are categorized into five risk levels: very low, low, moderate, high, and very high.

By visually comparing landslide susceptibility maps shown in Fig. 5 and Fig. 6, the SVM, XGBoost, CNN, and LSTM models consistently exhibited clearer delineation between landslide and non-landslide areas, with fewer regions falling into the 'moderate' category, irrespective of the factor sets employed. This observation aligned with the performance metrics presented in Table 5, which quantitatively confirm the superior classification accuracy of the aforementioned models. These advanced ML and DL models exhibited enhanced overall performance in predicting landslide-prone areas and assigning appropriate risk levels to critical regions, showcasing their potential for effective landslide susceptibility mapping. Their superior performance can be attributed to their ability to capture complex nonlinear relationships and interactions among the input factors, enabling them to model the intricate relations of landslide occurrence more effectively. In contrast, the LR model and statistical models (i.e., IV, FR, and WoE), exhibited lower discriminatory power between landslide and non-landslide areas in the generated LSM maps, with larger regions falling into the 'Moderate' category. This leads to suboptimal landslide susceptibility classifications, highlighting the limitations of these models in accurately identifying and delineating landslide-prone areas. These models rely on assumptions of linearity and independence among the input factors, which may not hold true in the context of landslide susceptibility assessment. Consequently, they may oversimplify the relationships between the contributing factors and landslide occurrence, leading to less accurate predictions and less informative susceptibility maps.

When comparing the maps generated using 19 contributing factors and 9 triggering factors, it becomes evident that the overall spatial patterns of landslide susceptibility remained consistent within each model. However, the maps derived from 9 triggering factors tended to have slightly more conservative risk level assignments, with a larger proportion of the study area classified as having low or moderate susceptibility. This observation aligned with the slightly lower values of the performance metrics obtained when using 9 triggering factors, as discussed in Section 4.1. The inclusion of a comprehensive set of contributing factors (19 factors) enhanced the models' ability to capture the complex interplay of various contributing factors influencing landslide occurrence, and therefore improved the overall model's prediction accuracy.

*4.3. Comparison of predicted susceptibility levels by different LSM models*

To further assess the performance of the LSM models, Fig. 7 and Fig. 8 illustrate the density of landslides across different landslide susceptibility levels predicted using various LSM models, corresponding to 19 contributing factors and 9 trigging factors, respectively. The values presented in Fig. 7 and Fig. 8 denote the percentage of landslides occurring within each susceptibility level. These values revealed the limited capabilities of the IV, FR, WoE, and LR models in delineating landslide susceptibility levels, regardless of the factor sets used. This was evidenced by the disproportionately high percentages predicted in the moderate, low, and high susceptibility classes. For instance, in the IV model with 19 factors (Fig. 7a), the percentage of landslides in the moderate and high susceptibility class was significantly higher than in the very low and very high classes. Similarly, the FR and WoE models (Fig. 7b and Fig. 7c) showed a high concentration of landslides in the high and moderate susceptibility classes. These findings suggest the inability of these models to effectively characterize the complex relationships and precisely demarcate landslide areas. In contrast, SVM, XGBoost, CNN, and LSTM exhibited enhanced predictive performance compared to conventional statistical methods (i.e., IV, FR, and WoE) and the linear ML model - LR. The lower percentages in the moderate susceptibility class indicated the competency of these models in determining complex correlations between input factors and landslide occurrences, thereby providing more accurate forecasts. For instance, XGBoost with 19 factors (Fig. 7f) exhibited the highest percentage of landslides in the very low susceptibility class while maintaining relatively lower percentages in other classes. Similarly, CNN and LSTM (Fig. 7g and Fig. 7h, respectively) displayed varying distributions of landslide percentages across different susceptibility levels, highlighting their ability to identify landslide susceptibility patterns effectively.

Comparing the distributions presented in Fig. 7 with those in Fig. 8 reveals that the models trained with 19 factors generally predicted a higher proportion of low susceptibility areas and a relatively lower proportion of high susceptibility areas compared to those trained with 9 factors. In contrast, the distributions shown in Fig. 8 were more balanced across all susceptibility levels, which may underestimate the actual probability of landslide occurrence in high-risk areas. The inclusion of a wider range of contributing factors (i.e. 19 factors) enabled models to more effectively capture the complex interplay of various landslide contributing factors, resulting in more accurate classification of susceptibility levels. This was particularly evident in the XGBoost, CNN, and LSTM models, where the percentage of landslides in the moderate susceptibility class was consistently lower when using 19 factors compared to 9 factors.

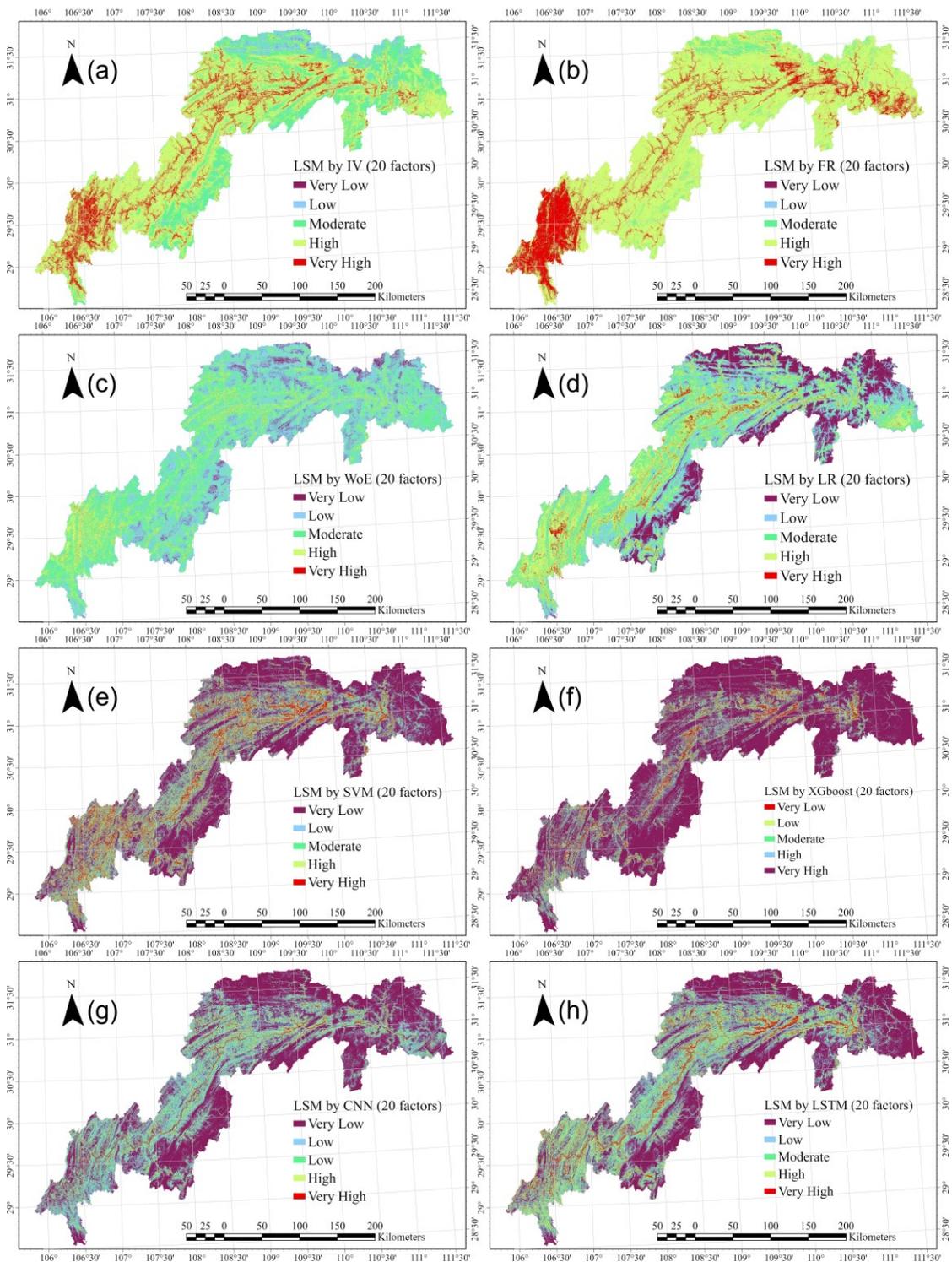

Fig. 5 Landslide susceptibility maps derived from 19 factors: (a) IV, (b) FR, (c) WoE, (d) LR, (e) SVM, (f) XGBoost, (g) CNN, (h) LSTM.

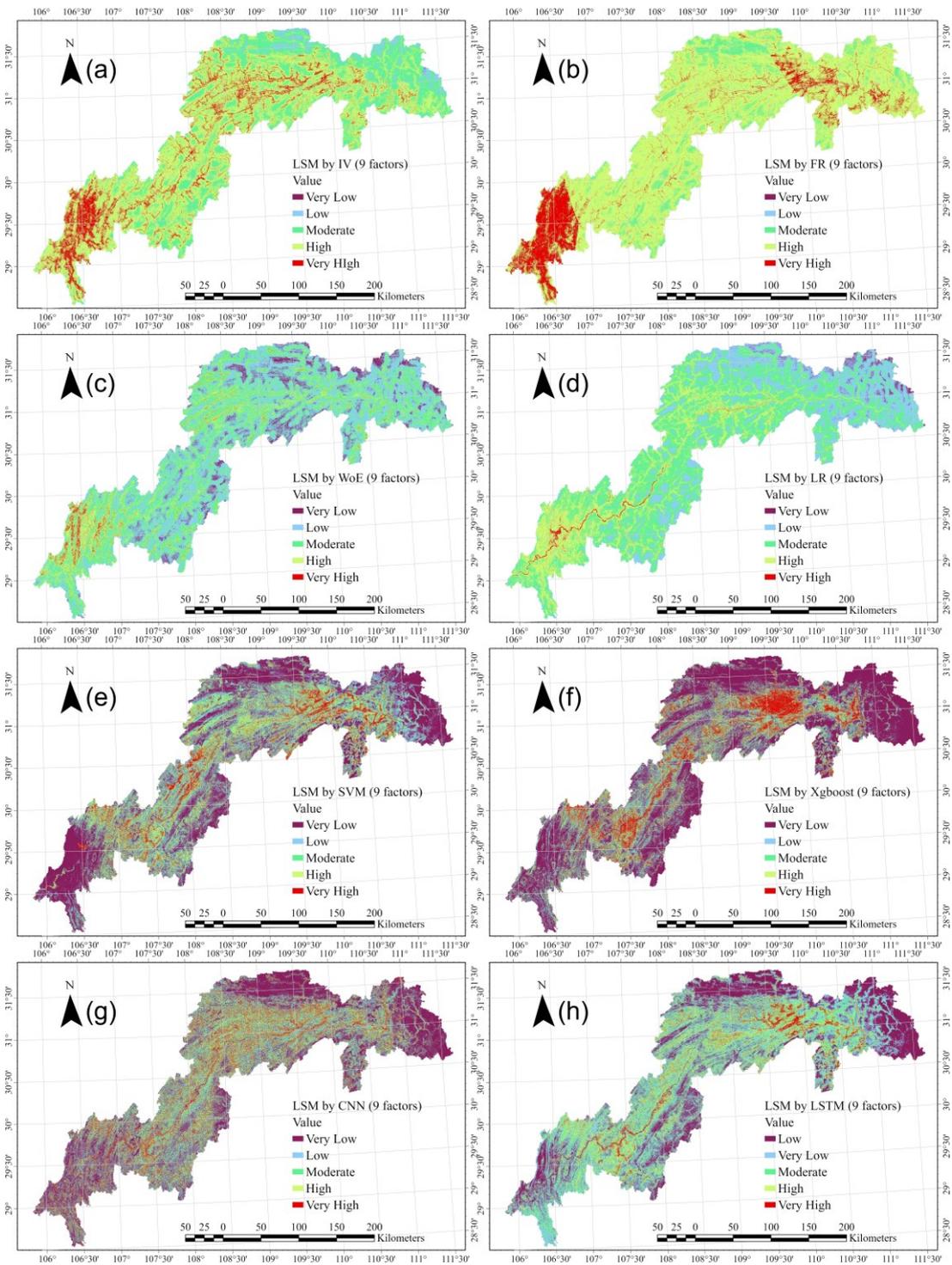

Fig. 6 Landslide susceptibility maps derived from 9 factors: (a) IV, (b) FR, (c) WoE, (d) LR, (e) SVM, (f) XGBoost, (g) CNN, (h) LSTM.

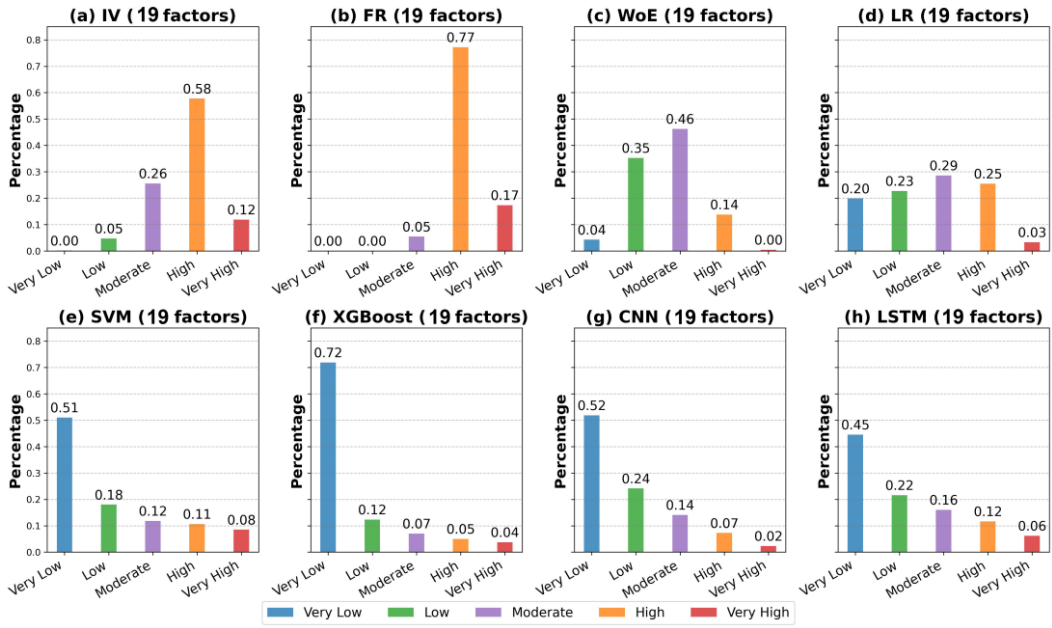

Fig. 7 Density of landslides with respect to the landslide susceptibility levels in different LSM methods considering 19 contributing factors: (a) IV, (b) FR, (c) WoE, (d) LR, (e) SVM, (f) XGBoost, (g) CNN, (h) LSTM.

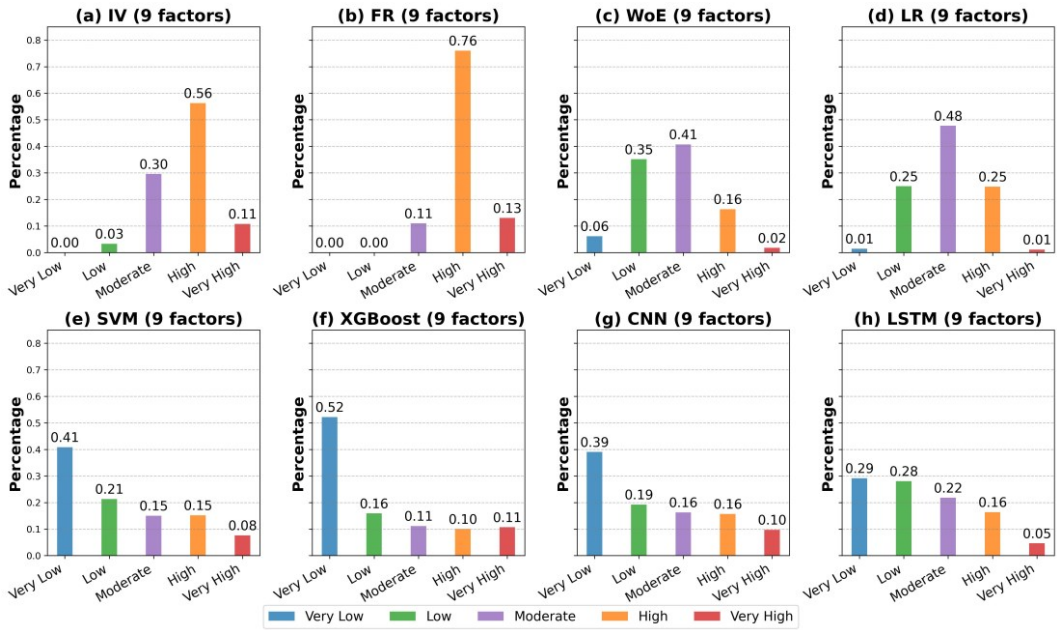

Fig. 8 Density of landslides with respect to the landslide susceptibility levels in different LSM methods considering 9 contributing factors: (a) IV, (b) FR, (c) WoE, (d) LR, (e) SVM, (f) XGBoost, (g) CNN, (h) LSTM.

*4.4. Interpretation Results*

4.4.1. Global interpretation through identified important factors by statistical analysis

To investigate the relationship between landslide occurrence and various contributing factors, this study conducted a comprehensive analysis using the Ordinary Least Squares (OLS) regression method. Two sets of regression analyses were performed, considering 19 contributing factors and 9 triggering factors, respectively. The coefficients of the regression model indicate both the strength and direction of the linear relationship between the factor variable and the binary landslide occurrence variable. Positive and negative coefficients suggest positive and negative correlations, respectively, between a factor and landslide occurrence. Larger absolute values of coefficients indicate a stronger influence. Standard errors reflect the precision of the coefficient estimates. Significance testing was conducted through t-statistics and p-values, which determines the statistical significance of the relationship between each factor and landslide occurrence. The t-statistic is calculated by dividing the coefficient by its standard error, and it measures the coefficient' deviation from zero in terms of standard deviations. A larger absolute t-statistic value indicates a more significant relationship. The p-value represents the probability of observing a t-statistic as extreme as the calculated one, under the assumption that the null hypothesis (i.e., no relationship between the factor and landslide occurrence) is true. A small p-value (typically $< 0.05$) indicates strong evidence against the null hypothesis, suggesting a statistically significant relationship between the factor and landslide occurrence.

Table 6 presents the OLS regression results for the case of 19 contributing factors. Among these, several factors demonstrated significant relationships with landslide occurrence. Elevation showed a significant negative correlation (coefficient = -0.0004, p-value $< 0.001$), implying that higher elevations were associated with a decreased likelihood of landslides. Conversely, slope exhibited a significant positive correlation (coefficient = 0.0281, p-value $< 0.001$), indicating that steeper slopes were more prone to landslide occurrence. Plan curvature also showed significant positive effects on landslide occurrence, with coefficients of 0.0334, and p-values $< 0.001$. The TRI and landform showed significant negative correlations with landslide occurrence, with coefficient values of -0.0719 and -0.0329, respectively, and p-values $< 0.001$, suggesting that areas with higher terrain ruggedness and certain landform characteristics may be less susceptible to landslides. Other factors, including cut depth, lithology, distance to fault, STI, TWI, distance to stream, land use, distance to road, peak rainfall intensity, and average rainfall intensity, also exhibited significant relationships with landslide occurrence, with p-values $< 0.001$, underscoring the importance of hydrological, vegetation, and rainfall factors in influencing landslide susceptibility. Factors such as aspect, profile curvature, SPI and NDVI did not show significant relationships with landslide occurrence, as their p-values exceeded the significance threshold of 0.05.

Table 1 OLS regression results for landslide occurrence considering 19 contributing factors.

|  | Coefficient | Standard Error | t-statistic | P-value |
|---|---|---|---|---|
| **Elevation** | -0.0004 | 5.66E-06 | -65.448 | $< 0.001$ |
| **Slope** | 0.0281 | 0.001 | 35.395 | $< 0.001$ |
| **Aspect** | -1.76E-05 | 1.72E-05 | -1.026 | 0.305 |
| **Plan Curvature** | 0.0334 | 0.003 | 12.12 | $< 0.001$ |
| **Profile Curvature** | -0.0013 | 0.002 | -0.566 | 0.571 |
| **Cut Depth** | 0.0026 | 0 | 12.113 | $< 0.001$ |
| **TRI** | -0.0719 | 0.002 | -29.881 | $< 0.001$ |
| **Landform** | -0.0329 | 0.002 | -14.572 | $< 0.001$ |
| **Lithology** | -0.0123 | 0.002 | -6.185 | $< 0.001$ |

| | Coefficient | Standard Error | t-statistic | P-value |
|---|---|---|---|---|
| **Soil Type** | 0.0463 | 0.002 | 23.352 | < 0.001 |
| **Distance to Fault** | -7.52E-07 | 1.13E-07 | -6.644 | < 0.001 |
| **STI** | -0.0039 | 0 | -12.247 | < 0.001 |
| **SPI** | -2.39E-06 | 6.35E-06 | -0.377 | 0.706 |
| **TWI** | 0.08 | 0.003 | 30.193 | < 0.001 |
| **Distance to Stream** | 3.30E-06 | 9.55E-07 | 3.46 | 0.001 |
| **NDVI** | -0.0051 | 0.013 | -0.4 | 0.689 |
| **Land Use** | 0.0388 | 0.001 | 41.325 | < 0.001 |
| **Distance to Road** | -2.18E-05 | 1.45E-06 | -14.983 | < 0.001 |
| **Peak Rainfall Intensity** | 0.012 | 0.001 | 16.354 | < 0.001 |
| **Average Rainfall Intensity** | 0.0947 | 0.007 | 13.215 | < 0.001 |

Table 7 presents the OLS regression results concerning 9 triggering factors. Among these factors, slope, lithology, TWI, distance to stream, NDVI, distance to road, peak rainfall intensity, and average rainfall intensity exhibited significant relationships with landslide occurrence, with p-values < 0.001. The significance of these factors in 9-factor scenario was consistent with their influence in the 19-factor scenario, indicating their robust influence on landslide occurrence. Distance to fault, however, did not show a significant relationship with landslide occurrence, as its p-value exceeded the significance threshold of 0.05.

Table 7 OLS regression results for landslide occurrence considering 9 triggering factors.

| | Coefficient | Standard Error | t-statistic | P-value |
|---|---|---|---|---|
| **Slope** | 0.0042 | 0 | 20.574 | < 0.001 |
| **Lithology** | 0.0177 | 0.002 | 10.615 | < 0.001 |
| **Distance to Fault** | 1.044E-07 | 1.2E-07 | 0.868 | 0.385 |
| **TWI** | 0.0433 | 0.002 | 26.238 | < 0.001 |
| **Distance to Stream** | -2.916E-05 | 9.16E-07 | -31.824 | < 0.001 |
| **NDVI** | -0.2072 | 0.013 | -16.19 | < 0.001 |
| **Distance to Road** | -5.912E-05 | 1.47E-06 | -40.314 | < 0.001 |
| **Peak Rainfall Intensity** | 0.0185 | 0.001 | 24.593 | < 0.001 |
| **Average Rainfall Intensity** | 0.1492 | 0.007 | 20.092 | < 0.001 |

In the 9-factor scenario, factors were selected according to the prior knowledge of landslide mechanisms. The significance of all factors except distance to fault in the 9-factor analysis confirmed their importance in trigging landslide occurrence. These trigging factors were considered to provide a physically meaningful and explicable approach to understanding landslide susceptibility. While the 19-factor analysis demonstrated higher prediction accuracy, as evident from the performance metrics in Table 5, some of the identified relationships might not align with the physical processes governing landslide occurrence. For example, the 19-factor analysis suggested that higher elevations were associated with a decreased likelihood of landslides, which was irrelevant to the established understanding of landslide processes. Such counterintuitive findings, despite their statistical significance, can lead to misinterpretations and potentially flawed decision-making in landslide risk management.

### 4.4.2. Global interpretation by IV, WoE, and FR

The global interpretability of IV, WoE, and FR models can be assessed by examining the weights assigned to each class within the contributing factors. Table A1 in Appendix A presents the detailed calculation processes and results for these models.

In the IV model, the IV value of each class indicates its contribution to landslide occurrence. Classes with higher positive IV values are more strongly associated with landslide occurrence, while those with lower negative values are less likely to be associated with landslides. For example, in the elevation factor, the class "-27 to 200" had the highest IV value of 0.6221, suggesting that this elevation range was most prone to landslides. Conversely, the class "1800<" exhibited the lowest IV value of -2.6657, indicating that elevations above 1800m were least likely to experience landslides. Similarly, for the slope factor, the class "15-20" showed the highest IV value of 0.3001, implying that slopes between 15° and 20° were most susceptible to landslides. In the surface cut depth factor, the class "20-40" had the highest IV value of 0.1937, indicating that cut depths between 20 and 40 m were most prone to landslides. The IV values for other factors, such as TRI, landform, lithology, STI, TWI, distance to stream, NDVI, land use, distance to road, and rainfall intensity, followed a similar pattern, with higher positive values indicating a stronger association with landslide occurrence and lower negative values suggesting a weaker association.

In the WoE model, the positive (W+) and negative (W-) weights reflect the spatial association between each class and landslide occurrence. Classes with higher positive W+ values and lower negative W- values are more strongly associated with landslide occurrence. For instance, in the landform factor, class "terraced hills" had the highest W+ value of 0.9741 and the lowest W- value of -0.0157, indicating that this landform type was most susceptible to landslides. Similarly, in the distance to road factor, the class "0-200" exhibited the highest W+ value of 0.6130 and the lowest W- value of -0.0919, suggesting that areas within 200 m of roads were most prone to landslides. The W+ and W- values for other factors followed a similar pattern, with higher positive W+ values and lower negative W- values indicating a stronger association with landslide occurrence.

In the FR model, the FR values represents the relative abundance of landslides within each class compared to the overall study area. Classes with FR values greater than 1 are more likely to be associated with landslides, while those with FR values less than 1 are less likely. For example, in the peak rainfall intensity factor, the class "22<" held the highest FR value of 1.4227, suggesting that areas with peak rainfall intensity above 22 mm/h were most prone to landslides. Similarly, in the elevation factor, the class "-27 – 200" possessed the highest FR value of 1.8628, indicating that elevations between -27 and 200 m were most susceptible to landslides. This pattern extended to other factors, with values greater than 1 indicating a higher likelihood of landslide occurrence and values less than 1 suggesting a lower likelihood.

By comparing the IV, WoE, and FR values across different factors and their respective classes, we can identify the most significant contributors to landslide susceptibility in the study area. Factors with consistently high positive IV and W+ values, low negative W- values, and FR values greater than 1 across multiple classes can be considered as strong indicators of landslide susceptibility. These factors included elevation, slope, surface cut depth, TRI, landform, lithology, STI, TWI, distance to stream, NDVI, land use, distance to road, and rainfall intensity. On the other hand, factors with inconsistent or weak associations across the three statistical models, such as aspect, plan and profile curvature, distance to faults, and SPI, might have a less significant influence on landslide occurrence in the study area.

### 4.4.3. Global interpretation by SHAP

Fig. 9 and Fig. 10 provide valuable insights into the importance of contributing factors and their impact on landslide susceptibility prediction across various ML and DL models using SHAP. The analysis revealed several key findings that highlighted the differences and similarities across different models when trained with either 19 contributing factors or 9 triggering factors.

When examining the scenario of 19 contributing factors (Fig. 9), consistently high SHAP values for terrain attributes such as elevation and slope across all models underscored their fundamental role in characterizing landslide occurrence. This finding aligned well with the established understanding that topographic factors significantly influenced the prediction of landslide occurrence. However, the negligible influence of curvature-related factors, like profile curvature, suggested that these factors might not contribute substantially to the models' predictive power, possibly due to their localized nature or limited representation of the overall terrain morphology. Interestingly, the relative importance of factors varied among different models, indicating that each model captured and leveraged the information provided by contributing factors differently. For example, the prominence of slope in the LR model and SPI in the SVM model highlighted their unique capabilities in extracting relevant information from these specific factors. Moreover, the varying correlation directions of certain factors across models suggested that the relationship between these factors and landslide susceptibility might be complex and nonlinear, with different models capturing distinct aspects of these relationships.

In the scenario of 9 triggering factors (Fig. 10), the pattern of SHAP values showed greater consistency across different models compared to the 19-factor scenario. This suggested that various considered models consistently captured the relationships between these triggering factors and landslide occurrence. For example, slope and rainfall intensity (both peak and average) were identified as key trigging factors consistently across different models, aligning with engineering expectations that geomorphological and meteorological conditions significantly influence slope instability and landslide initiation. The variability in SHAP values for distance to road and NDVI suggested that their relationships with landslide occurrence might be more complex and harder to capture consistently across different models. This variability could be attributed to the presence of interactions, nonlinear relationships, or a weaker overall influence on landslide susceptibility compared to other factors. Different models may capture these complex relationships to varying degrees, leading to slight variations in their importance rankings. This finding emphasized the need for careful consideration and interpretation of these factors, as their influence on landslide susceptibility may be moderated by other variables or may manifest differently across different landscapes or environmental conditions.

The SHAP summary plots in Fig. 9 and Fig.10 provide a visually intuitive way to interpret factors' importance and their impact on landslide susceptibility. The position of each factor on the vertical axis indicates its overall importance (descending from top to bottom), while the spread and distribution of SHAP values along the horizontal axis reveal the range and directionality of its influence. The color gradient adds an additional layer of information, with red indicating high factor values and blue representing low values, allowing for a more nuanced understanding of how the factors' magnitudes contribute to the models' predictions.

This understanding becomes particularly relevant when comparing the SHAP summary plots across different models and input factor scenarios. In simpler, linear models like LR and SVM (Fig. 9a, b, c, d), the SHAP values exhibited a more clear-cut interpretation, with distinct clusters of red and blue points indicating the positive and negative influences of the factors, respectively. This suggested that these models captured the relationships between factors and landslide susceptibility in a more straightforward manner, aligning with the expected directionality of their effects. However, as the complexity of the models increased, as seen in XGBoost, CNN, and LSTM (Fig. 9 e, f, g, h, i, j), the SHAP summary plots revealed more complex patterns and distributions of SHAP values. The color gradients became more intermingled, and the spread of points along the horizontal axis widened, indicating that these models captured more complex, nonlinear relationships between factors and landslide susceptibility. While this increased complexity may lead to higher predictive accuracy, it can also introduce a certain level of deviation from expected interpretability, as models may identify subtle interactions and patterns that were not immediately apparent from a domain knowledge perspective.

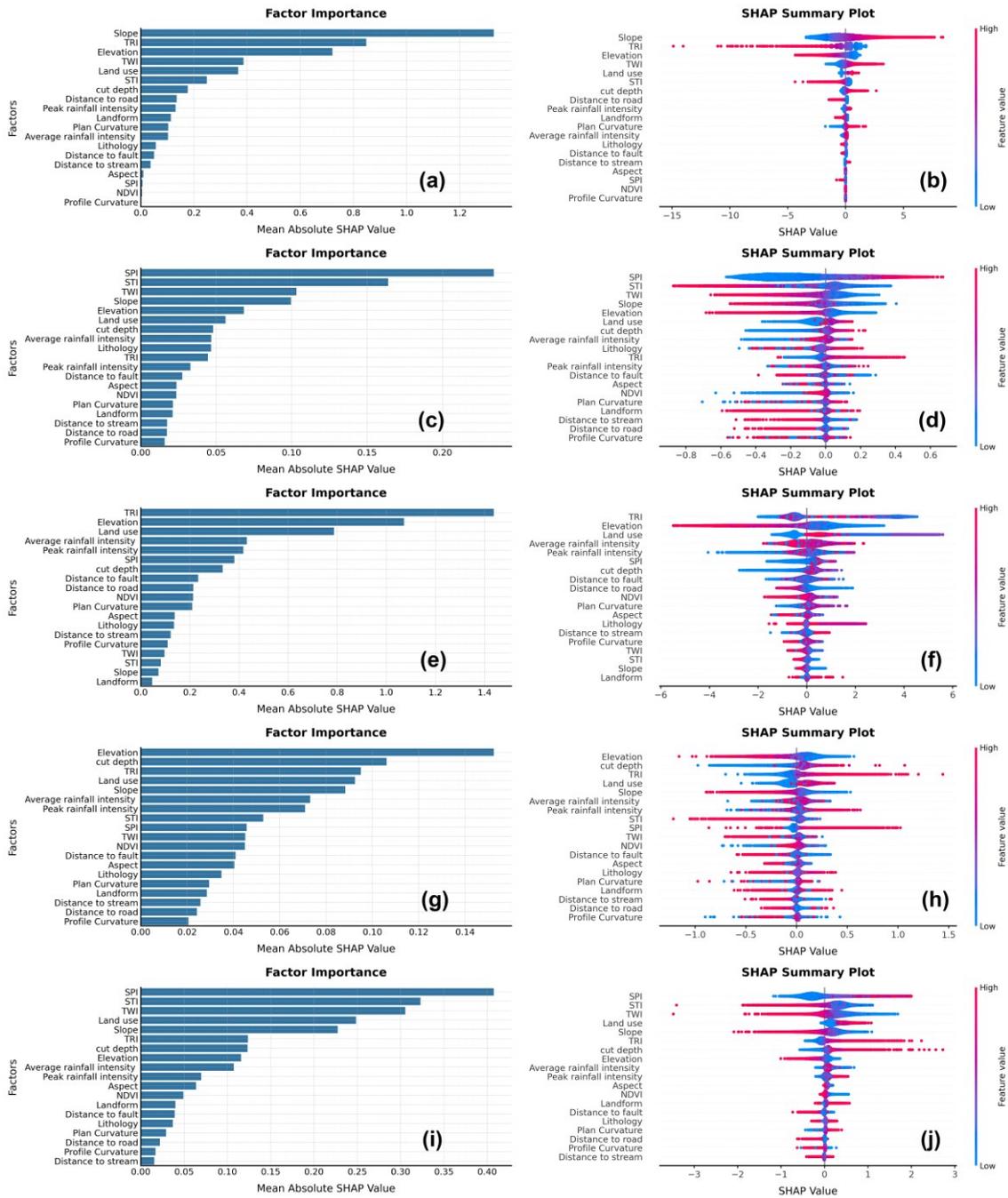

Fig. 9. Factor importance and SHAP values for LSM models concerning 19 contributing factors: (a) Factor importance in the LR model. (b) SHAP values in the LR model. (c) Factor importance in the SVM model. (d) SHAP values in the SVM model. (e) Factor importance in the XGBoost model. (f) SHAP values in the XGBoost model. (g) Factor importance in the CNN model. (h) SHAP values in the CNN model. (i) Factor importance in the LSTM model. (j) SHAP values in the LSTM model.

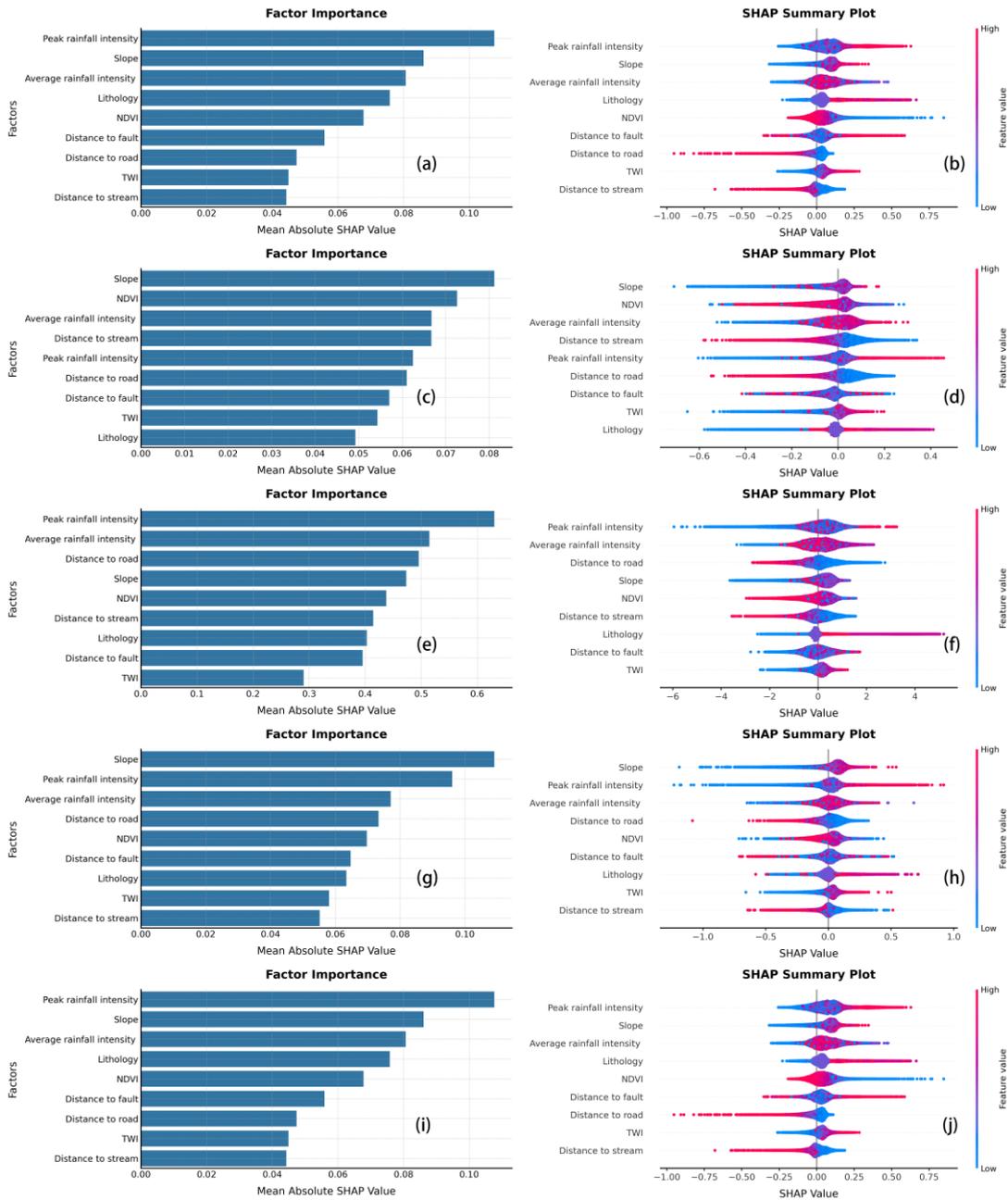

Fig. 10. Factor importance and SHAP values for LSM models concerning 9 contributing factors: (a) Factor importance in the LR model. (b) SHAP values in the LR model. (c) Factor importance in the SVM model. (d) SHAP values in the SVM model. (e) Factor importance in the XGBoost model. (f) SHAP values in the XGBoost model. (g) Factor importance in the CNN model. (h) SHAP values in the CNN model. (i) Factor importance in the LSTM model. (j) SHAP values in the LSTM model.

### 4.4.4. Global interpretation by DeepLIFT

The DeepLIFT analysis results, as depicted in Fig.11 and Fig.12, shed light on the significance of contributing factors in landslide susceptibility prediction using CNN and LSTM models. The attribution patterns varied with the scenarios of input factors and model architectures. In the scenario of 19 contributing factors (Fig.11), the CNN model emphasized the importance of topographic indices, with TRI and TWI ranking highest in factor importance, followed by land use and rainfall intensity. The LSTM model, in contrast, highlighted the significance of SPI, TWI and STI, along with cut depth, land use, and rainfall intensity. In the scenario of 9 trigging factors (Fig.12), both CNN and LSTM models exhibited a more aligned attribution pattern. Specifically, lithology and rainfall intensity (both average and peak) emerged as the most important factors, while TWI and NDVI were consistently identified as least influential factors, according to both CNN and LSTM models

The DeepLIFT summary plots offer a visually intuitive representation of factor importance and their impact on the models' predictions. Comparing the DeepLIFT results between the 19-facotr and 9-factor scenarios revealed the intricate interaction of factors and the varying attribution patterns based on input factors and model architectures. The scenario of 19 contributing factors highlighted more complex interactions among factors, with topographic indices playing a crucial role, particularly in the LSTM model. In contrast, the scenario of 9 trigging factors underscored the importance of lithology and rainfall intensity, with consistent rankings and attribution patterns in both CNN and LSTM models.

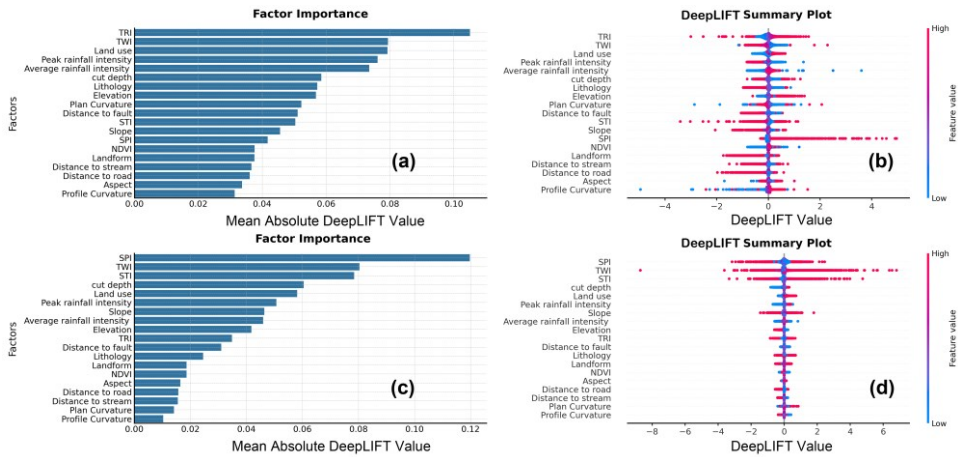

Fig. 11. Importance of contributing factors in DL-based LSM models by DeepLIFT: (a) CNN, (d) LSTM.

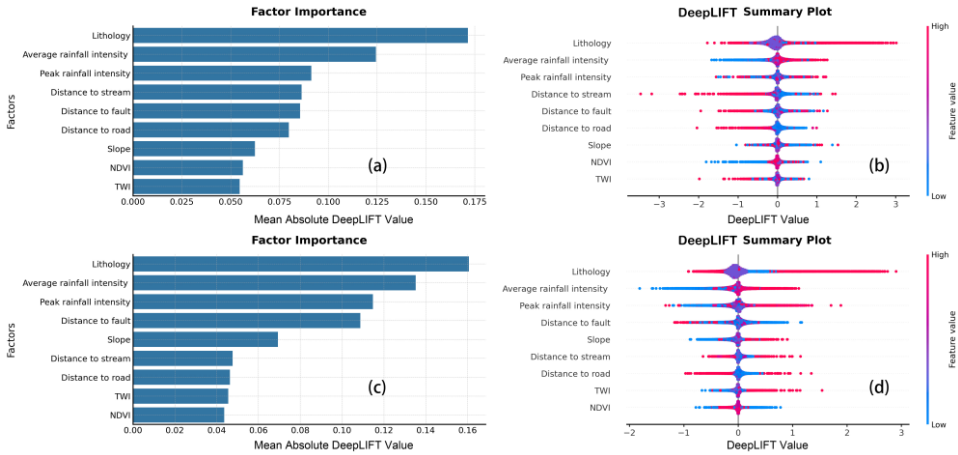

Fig. 12. Importance of contributing factors in DL-based LSM models by DeepLIFT: (a) CNN, (d) LSTM.

4.4.5. Local interpretation

The Wushanping landslide (Chen and Fan 2023a) was selected for studying local interpretation consistency. Table 8 lists the primary contributing factors ranked by importance scores, which were identified by each combination of LSM models and interpretability methods considered in this study. These factors were compared with the actual trigging factors documented in the field investigation report for the Wushanping landslide.

Using 19 contributing factors, different combinations of LSM models and interpretability methods exhibited variations in identifying the key factors influencing the Wushanping landslide. The combination of LR model with LIME, SVM model with SHAP, and XGBoost model with LIME emphasized topographical, land use, and geological factors such as elevation, slope, lithology, and landform. In contrast, the combination of CNN model with DeepLIFT highlighted the importance of slope, peak rainfall intensity, and plan curvature, which are more directly related to the factors mentioned in the field investigation report. When comparing the interpreted results of different LSM-interpretability model combinations with the field investigation report, we observed that most combinations successfully captured slope as a critical factor. However, only the combination of CNN model with DeepLIFT identified peak rainfall intensity, which was one of the key factors reported in the field investigation report. Other factors such as lithology and average rainfall intensity were not consistently recognized across all LSM-interpretability model combinations when using 19 contributing factors.

When using 9 triggering factors (as shown in Table 8), the local interpretation results exhibited a consistent capture of the core factors driving the Wushanping landslide, despite slight variations across different combinations of LSM models and interpretation methods. Slope and rainfall intensity (either peak or average) were consistently identified as dominant factors by most combinations of LSM models and interpretation methods, aligning with the findings of the field investigation report. This highlighted the models' ability to capture critical factors when trained with factors directly associated with landslide trigging mechanisms. The inclusion of distance to fault and TWI in certain cases suggested the potential influence of geological structure and soil moisture conditions on landslide occurrence.

For local interpretation, we observed factors relevant to the actual causes of the landslide, such as slope, lithology, and rainfall intensity. This indicated that LSM models, combined with suitable interpretation methods, could reflect critical factors affecting landslide occurrence. Meanwhile, certain key factors identified by each combination of LSM model and interpretability method varied, likely influenced by model architectures and training data. The results demonstrated the potential of interpretable LSM models in reflecting important landslide characteristics, while also showing that current data-driven models combined with interpretation methods may not fully explain the causal relationships between landslides and related factors. The use of a more dedicated set of triggering factors (9 factors) led to more consistent and accurate identification of the key factors driving the Wushanping landslide compared to the broader set of contributing factors (19 factors).

Table.8 Local interpretation of landslide-related factors identified by different interpretation methods for various LSM models in the Wushanping landslide.

| Input | Interpretation Method | LR | SVM | XGBoost | LSTM | CNN |
|---|---|---|---|---|---|---|
| **19 factors** | **LIME** | Elevation, Slope, Lithology | Peak rainfall intensity, Slope, Distance to fault | Landform, Lithology, Slope | Peak rainfall intensity, Elevation, Distance to stream | Elevation, Landform, NDVI |
| | **SHAP** | Slope, Aspect, Average rainfall intensity | Slope, Land use, Elevation | Cut depth, NDVI | Elevation, Peak rainfall intensity, Landform | Slope, STI, Distance to road |
| | **DeepLIFT** | - | - | - | Elevation, Slope, Average rainfall intensity | Slope, Peak rainfall intensity, Plan Curvature |
| **9 factors** | **LIME** | Slope, Lithology, Peak rainfall intensity | Slope, Distance to fault, Average rainfall intensity | Slope, TWI, Distance to road | Slope, Peak rainfall intensity, NDVI | Slope, Lithology, Distance to stream |
| | **SHAP** | Slope, Average rainfall intensity, Lithology | Slope, Peak rainfall intensity, TWI | Slope, Lithology, NDVI | Slope, Distance to fault, Average rainfall intensity | Slope, Peak rainfall intensity, Distance to road |
| | **DeepLIFT** | - | - | - | Slope, Lithology, Distance to stream | Slope, Average rainfall intensity, TWI |
| | **Field investigation reports (Chen and Fan 2023a)** | Lithology, Slope, Peak rainfall intensity, Average rainfall intensity | | | | |

## 5. Discussion

### 5.1. Distinguishing correlation and causation

Data-driven models are adept at finding complex correlations between landslide contributing factors and landslide occurrence. However, these models only identify correlations, but do not necessarily reveal true causal mechanisms. For instance, heavy rainfall may have a strong correlation with landslides, but it may not be the sole cause of landslides, as other factors like geological structure and human activities may also be involved. Therefore, performing effective LSM requires more than just data-driven models. It is crucial to combine expertise from geology, hydrology, engineering, and other domains, as well as field surveys, to understand and validate the underlying causal relationships behind predicted correlations.

The comparative analysis of models trained with 19 contributing factors and 9 triggering factors highlights the importance of distinguishing between correlation and causation. When using a broader set of contributing factors, models may identify factors that are correlated with landslide occurrence but in reality do not contribute to the formation of landslides. This can lead to inconsistent factor importance rankings and interpretations across different models. In contrast, focusing on a specific set of triggering factors that are more directly linked to landslide causation can improve the consistency and accuracy of model interpretations, as evidenced by the results obtained using 9 triggering factors.

To bridge the gap between correlation and causation, future research could explore the integration of causal inference techniques with data-driven models. By incorporating domain knowledge and causal assumptions into the modeling process, we can move beyond purely data-driven correlations and develop models that better capture the underlying causal mechanisms of landslides. This may involve the use of causal graphs, structural equation modeling, or counterfactual reasoning to explicitly represent and reason about the causal relationships between factors and landslide occurrence.

*5.2. The relationship between accuracy, interpretability, and reliability of LSM models*

In the field of LSM, ML and DL models have gained attention due to their ability to significantly improve prediction accuracy. However, compared to the intuitive interpretability of traditional statistical models, these advanced models often lack intuitive interpretability due to their "black box" nature. Such uncertainty in interpretability may affect the credibility and practical value of model predictions. Therefore, when applying complex models for landslide prediction, it is important to appreciate limitations in interpretability and improve result reliability by combining expert assessment and field validation. Existing data-driven interpretable models still have limitations and cannot robustly and accurately obtain the real driving factors of landslides. Building a more interpretable and robust framework for landslide prediction remains an important challenge in current landslide prediction research. The comparative analysis of models trained with different sets of factors (19 contributing factors vs. 9 triggering factors) sheds light on the trade-off between accuracy and interpretability. While using a more comprehensive set of factors (e.g., 19 contributing factors in this study) may lead to higher prediction accuracy, it can also introduce complexities and inconsistencies in model interpretations. On the other hand, focusing on a dedicated set of triggering factors (e.g., 9 triggering factors in this study) may sacrifice some predictive accuracy but can improve the interpretability and reliability of models' predictions.

*5.3. Future work*

Future research in LSM should prioritize the development of models that better capture the complex causal mechanisms underlying landslide events. As demonstrated in this study, relying on a single model or interpretation method is insufficient to fully understand the intricate causation of landslides. To address this challenge, integrating causal inference techniques, such as causal graphs, structural equation modeling, or counterfactual reasoning, with data-driven models offers a promising approach. By explicitly representing and reasoning the causal relationships between contributing factors and landslide occurrence, such integrated models can provide a more comprehensive understanding of the landslide phenomenon. However, it should be noted that the effectiveness of such integrations may be affected by factors such as the quality and completeness of the causal knowledge incorporated, the compatibility between causal inference techniques and the data-driven models, and the interpretability of integrated models. These potential limitations should be carefully considered and addressed in future research.

In addition to causal inference, the development of inherently interpretable models should be explored. Rule-based systems or other transparent model architectures that maintain high prediction accuracy while allowing for clear interpretation of the decision-making process can enhance the interpretability of LSM results. Incorporating domain knowledge and physical constraints into the model training process is another crucial aspect of guiding models towards more realistic and interpretable solutions. This can be achieved by leveraging prior knowledge, expert rules, or physically-based constraints to regularize the model learning process. However, it is essential to strike a balance between the incorporation of domain knowledge and the flexibility of models to learn from data, as overly restrictive constraints may limit the models' ability to capture novel patterns or relationships.

To validate model interpretations and improve their reliability, conducting comprehensive field investigations and collecting high-quality data on landslide events is essential. Such data should include detailed information on the geological, hydrological, and environmental conditions surrounding occurred landslides. By incorporating diverse data types from multi-modal sources, such as remote sensing imagery and geophysical data models can capture a wider range of factors influencing landslide occurrence and potentially uncover new insights into underlying causal mechanisms. However, integrating multi-modal data may introduce additional challenges, such as data inconsistency, noise, and the need for effective data fusion techniques. Future research should investigate robust methods for handling these challenges and ensuring the quality of integrated data.

Developing hybrid models that combine data-driven approaches with physically-based models or expert systems represents a great opportunity to enhance LSM by leveraging the strengths of different models. Hybrid models could integrate the predictive power of data-driven models with the interpretability and domain knowledge of physically-based models or expert systems, potentially leading to more accurate and interpretable LSM results. However, the success of hybrid models depends on an effective integration of different model components, which may require careful design and validation to ensure compatibility and avoid potential contradictions or inconsistencies in future research.

To assess the robustness and consistency of different LSM models and interpretation methods, it is crucial to conduct systematic model intercomparison across various study areas and datasets. This can help identify the strengths and limitations of each approach and guide the selection of appropriate models for specific contexts. Furthermore, ensemble analysis techniques could be employed to combine the predictions and interpretations of multiple models, potentially improving overall performance and reliability. However, the effectiveness of ensemble analysis may be influenced by factors such as the diversity and quality of the individual models, the appropriateness of the combination strategy, and the interpretability of the resulting ensemble model. Future research should investigate methods for optimizing ensemble analysis in the context of LSM and developing techniques for interpreting the results of ensemble models.

The aforementioned research directions can contribute to the development of more accurate, interpretable, and reliable LSM models that better align with the underlying causal mechanisms of landslides. This, in turn, will support the establishment of more effective landslide risk assessment and management strategies, ultimately helping mitigate the impacts of devastating landslides on communities and infrastructure. As the research on LSM continues to evolve, it is essential for researchers to collaborate across disciplines, share knowledge, and leverage advancements in data collection, modeling techniques, and interpretation methods to improve our understanding and prediction of landslide events. However, it should be acknowledged that the success of future research on LSM will depend on the availability of high-quality data, the effectiveness of proposed methodologies, and the ability to communicate findings to stakeholders and decision-makers in a clear and actionable manner.

## 6. Conclusion

This study evaluated the interpretability and accuracy of statistical, ML and DL models in predicting landslide susceptibility. Experiments were conducted using two sets of factors: 19 contributing factors and 9 triggering factors. The results showed that the CNN model achieved the highest prediction accuracy of 0.8447 when using 19 contributing factors and accuracy of 0.8048 when using 9 triggering factors, significantly outperforming traditional statistical models. Additionally, advanced ML models like XGBoost and SVM also demonstrated strong predictive capabilities, with accuracy rates of 0.8420 and 0.8364, respectively, when using 19 factors, and 0.8177 and 0.7837, respectively, when using 9 factors. These results indicate that DL and sophisticated ML algorithms can more accurately establish complex nonlinear relationships between input factors and landslide occurrence, effectively capturing the intricate interplay of various conditions, including terrain, geology, hydrology, soil, and vegetation, in determining landslide susceptibility.

However, in terms of the interpretability of predictions, this study discovered that different types of models, coupled with interpretation methods, exhibited inconsistent understandings and identifications of key landslide influencing factors. Statistical, ML and DL models exhibited distinctions in assessing the importance of factors, particularly when using a broader set of contributing factors (19 factors). This discrepancy may originate from differences in the patterns learned from the dataset by each model, as well as their formed perceptions about the mechanisms behind landslide occurrences. Moreover, utilizing explanation methods like SHAP, LIME, and DeepLIFT on the same model also led to variations in interpreted key factors.

The comparative analysis of the outputs from models trained with 19 contributing factors and 9 triggering factors highlighted the trade-off between accuracy and interpretability. While using a comprehensive set of contributing factors led to higher prediction accuracy, it also introduced complexities and inconsistencies in model interpretations. On the other hand, focusing on a dedicated set of triggering factors sacrificed some predictive power but improved the interpretability of the models, as evidenced by more consistent key trigging factors identified across different models and the alignment of local interpretations with field investigation reports.
.

## Authorship contribution statement

C.C. and L.F. participated in the conceptualization and design of the study. C.C. conducted data analysis, coding, and contributed to the writing of the original draft. L.F. provided oversight and contributed to the review process.

## Declaration of Competing Interest

This research was funded by Xi'an Jiaotong-Liverpool University Research Enhancement Fund under grant number REF-21-01-003.This research was funded by Xi'an Jiaotong-Liverpool University Research Enhancement Fund under grant number REF-21-01-003.

## Data availability

Data will be made available on request.

modeling. *Applied Soft Computing, 142*, 110324

Reichenbach, P., Rossi, M., Malamud, B.D., Mihir, M., & Guzzetti, F. (2018). A review of statistically-based landslide susceptibility models. *Earth-science reviews, 180*, 60-91

Russell, M., Kershaw, J., Xia, Y., Lv, T., Li, Y., Ghassemi-Armaki, H., Carlson, B.E., & Wang, P. (2023). Comparison and explanation of data-driven modeling for weld quality prediction in resistance spot welding. *Journal of Intelligent Manufacturing*, 1-15

Saito, H., Nakayama, D., & Matsuyama, H. (2009). Comparison of landslide susceptibility based on a decision-tree model and actual landslide occurrence: the Akaishi Mountains, Japan. *Geomorphology, 109*, 108-121

Sun, D., Chen, D., Zhang, J., Mi, C., Gu, Q., & Wen, H. (2023). Landslide Susceptibility Mapping Based on Interpretable Machine Learning from the Perspective of Geomorphological Differentiation. *Land, 12*, 1018

Sun, D., Ding, Y., Zhang, J., Wen, H., Wang, Y., Xu, J., Zhou, X., & Liu, R. (2022). Essential insights into decision mechanism of landslide susceptibility mapping based on different machine learning models. *Geocarto International*, 1-29

Sun, D., Xu, J., Wen, H., & Wang, D. (2021). Assessment of landslide susceptibility mapping based on Bayesian hyperparameter optimization: A comparison between logistic regression and random forest. *Engineering Geology, 281*, 105972

Tang, H., Wasowski, J., & Juang, C.H. (2019). Geohazards in the three Gorges Reservoir Area, China – Lessons learned from decades of research. *Engineering Geology, 261*, 105267

Van Westen, C.J. (1993). Application of geographic information systems to landslide hazard zonation

Wang, Q., Guo, Y., Li, W., He, J., & Wu, Z. (2019). Predictive modeling of landslide hazards in Wen County, northwestern China based on information value, weights-of-evidence, and certainty factor. *Geomatics, Natural Hazards and Risk, 10*, 820-835

Xiao, T., Segoni, S., Liang, X., Yin, K., & Casagli, N. (2023). Generating soil thickness maps by means of geomorphological-empirical approach and random forest algorithm in Wanzhou County, Three Gorges Reservoir. *Geoscience Frontiers, 14*, 101514

Xiao, T., Yin, K., Yao, T., & Liu, S. (2019). Spatial prediction of landslide susceptibility using GIS-based statistical and machine learning models in Wanzhou County, Three Gorges Reservoir, China. *Acta Geochimica, 38*, 654-669

Yao, W., Li, C., Zuo, Q., Zhan, H., & Criss, R.E. (2019). Spatiotemporal deformation characteristics and triggering factors of Baijiabao landslide in Three Gorges Reservoir region, China. *Geomorphology, 343*, 34-47

Yilmaz, I. (2009). Landslide susceptibility mapping using frequency ratio, logistic regression, artificial neural networks and their comparison: a case study from Kat landslides (Tokat—Turkey). *Computers & Geosciences, 35*, 1125-1138

Yin, K. (1988). Statistical prediction model for slope instability of metamorphosed rocks. In, *Proceedings of the 5th International Symposium on Landslides* (pp. 1269-1272)

Zêzere, J., Pereira, S., Melo, R., Oliveira, S., & Garcia, R.A. (2017). Mapping landslide susceptibility using data-driven methods. *Science of the total environment, 589*, 250-267

Zhang, J., Ma, X., Zhang, J., Sun, D., Zhou, X., Mi, C., & Wen, H. (2023). Insights into geospatial heterogeneity of landslide susceptibility based on the SHAP-XGBoost model. *Journal of Environmental Management, 332*, 117357

Zhang, K., Wu, X., Niu, R., Yang, K., & Zhao, L. (2017). The assessment of landslide susceptibility mapping using random forest and decision tree methods in the Three Gorges Reservoir area, China. *Environmental Earth Sciences, 76*, 1-20

Zhou, X., Wen, H., Li, Z., Zhang, H., & Zhang, W. (2022). An interpretable model for the susceptibility of rainfall-induced shallow landslides based on SHAP and XGBoost. *Geocarto International, 37*, 13419-13450

Zhou, X., Wen, H., Zhang, Y., Xu, J., & Zhang, W. (2021). Landslide susceptibility mapping using hybrid random forest with GeoDetector and RFE for factor optimization. *Geoscience Frontiers, 12*, 101211
31

32Appendix A: Table A1 Results of WoE, FR, and IV

| Factor | Class | No. of total pixels in domain | No. of landslide pixels in domain | FR | IV | W+ | W− | C | S(c) | WOE |
|---|---|---|---|---|---|---|---|---|---|---|
| Elevation (m) | -27 – 200 | 27123070 | 1501 | 1.8628 | 0.6221 | 0.6221 | -0.0720 | 0.6941 | 0.0278 | 24.9535 |
| | 200–400 | 80664423 | 3801 | 1.5861 | 0.4613 | 0.4613 | -0.1827 | 0.6440 | 0.0201 | 31.9669 |
| | 400–600 | 65650313 | 2282 | 1.1700 | 0.1570 | 0.1570 | -0.0382 | 0.1952 | 0.0236 | 8.2810 |
| | 600–800 | 52467896 | 1528 | 0.9803 | -0.0199 | -0.0199 | 0.0033 | -0.0232 | 0.0276 | -0.8420 |
| | 800–1000 | 42920752 | 937 | 0.7348 | -0.3081 | -0.3081 | 0.0349 | -0.3430 | 0.0342 | -10.0328 |
| | 1000–1200 | 36429219 | 439 | 0.4056 | -0.9023 | -0.9023 | 0.0640 | -0.9664 | 0.0487 | -19.8318 |
| | 1200–1400 | 28535705 | 201 | 0.2371 | -1.4393 | -1.4393 | 0.0629 | -1.5022 | 0.0712 | -21.0983 |
| | 1400–1600 | 19319849 | 84 | 0.1463 | -1.9218 | -1.9218 | 0.0468 | -1.9685 | 0.1095 | -17.9715 |
| | 1600–1800 | 10721160 | 36 | 0.1130 | -2.1801 | -2.1801 | 0.0266 | -2.2067 | 0.1669 | -13.2182 |
| | 1800< | 9195561 | 19 | 0.0695 | -2.6657 | -2.6657 | 0.0238 | -2.6896 | 0.2296 | -11.7132 |
| Slope (°) | 0–5 | 27546473 | 400 | 0.4903 | -0.7128 | -0.7128 | 0.0415 | -0.7542 | 0.0510 | -14.7983 |
| | 5–10 | 47474301 | 1143 | 0.8129 | -0.2071 | -0.2071 | 0.0281 | -0.2352 | 0.0313 | -7.5133 |
| | 10–15 | 54961697 | 1951 | 1.1985 | 0.1811 | 0.1811 | -0.0365 | 0.2176 | 0.0251 | 8.6865 |
| | 15–20 | 61053290 | 2445 | 1.3521 | 0.3017 | 0.3017 | -0.0748 | 0.3765 | 0.0230 | 16.3395 |
| | 20–25 | 54098287 | 1960 | 1.2233 | 0.2015 | 0.2015 | -0.0404 | 0.2419 | 0.0250 | 9.6739 |
| | 25–30 | 46424437 | 1340 | 0.9746 | -0.0258 | -0.0258 | 0.0038 | -0.0295 | 0.0292 | -1.0105 |
| | 30–35 | 33465994 | 763 | 0.7698 | -0.2616 | -0.2616 | 0.0234 | -0.2850 | 0.0376 | -7.5849 |
| | 35–40 | 21534544 | 418 | 0.6554 | -0.4225 | -0.4225 | 0.0217 | -0.4443 | 0.0499 | -8.9029 |
| | 40–45 | 12623036 | 218 | 0.5831 | -0.5394 | -0.5394 | 0.0151 | -0.5545 | 0.0684 | -8.1021 |
| | 45< | 13630126 | 190 | 0.4707 | -0.7536 | -0.7536 | 0.0207 | -0.7743 | 0.0732 | -10.5769 |
| Aspect | Flat | 2791707 | 24 | 0.2944 | -1.2227 | -1.2227 | 0.0061 | -1.2288 | 0.2044 | -6.0122 |
| | North | 43412973 | 1281 | 1.0106 | 0.0106 | 0.0106 | -0.0017 | 0.0122 | 0.0301 | 0.4067 |
| | North_east | 40847454 | 1128 | 0.9458 | -0.0557 | -0.0557 | 0.0078 | -0.0635 | 0.0317 | -2.0021 |
| | East | 47567258 | 1384 | 0.9965 | -0.0035 | -0.0035 | 0.0006 | -0.0041 | 0.0291 | -0.1402 |
| | South_east | 49477073 | 1442 | 0.9982 | -0.0018 | -0.0018 | 0.0003 | -0.0021 | 0.0286 | -0.0739 |
| | South | 46012253 | 1337 | 0.9952 | -0.0048 | -0.0048 | 0.0008 | -0.0056 | 0.0295 | -0.1892 |
| | South_west | 45356892 | 1394 | 1.0526 | 0.0513 | 0.0513 | -0.0087 | 0.0600 | 0.0290 | 2.0662 |
| | West | 47032786 | 1426 | 1.0384 | 0.0377 | 0.0377 | -0.0066 | 0.0443 | 0.0287 | 1.5410 |
| | North_west | 50313789 | 1412 | 0.9612 | -0.0396 | -0.0396 | 0.0071 | -0.0467 | 0.0289 | -1.6190 |
| Plan Curvature | Convex | 128067335 | 3626 | 0.9715 | -0.0289 | -0.0289 | 0.0331 | -0.0620 | 0.0241 | -2.5707 |
| | Flat | 108280957 | 3262 | 1.0337 | 0.0331 | 0.0331 | -0.0289 | 0.0620 | 0.0241 | 2.5707 |
| | Concave | 136679655 | 3939 | 0.9889 | -0.0112 | -0.0112 | 0.0151 | -0.0263 | 0.0244 | -1.0812 |
| Profile Curvature | Convex | 131083081 | 3667 | 0.9531 | -0.0481 | -0.0481 | 0.0582 | -0.1062 | 0.0242 | -4.3902 |
| | Flat | 102731296 | 3196 | 1.0599 | 0.0582 | 0.0582 | -0.0481 | 0.1062 | 0.0242 | 4.3902 |
| | Concave | 139213570 | 3964 | 0.9701 | -0.0304 | -0.0304 | 0.0431 | -0.0735 | 0.0244 | -3.0057 |
| Surface cut depth | 0–20 | 145547043 | 4112 | 0.9620 | -0.0387 | -0.0387 | 0.0247 | -0.0634 | 0.0198 | -3.1970 |
| | 20–40 | 149760126 | 5338 | 1.2137 | 0.1937 | 0.1937 | -0.1599 | 0.3536 | 0.0193 | 18.3384 |
| | 40–60 | 56244410 | 1123 | 0.6799 | -0.3858 | -0.3858 | 0.0564 | -0.4423 | 0.0315 | -14.0257 |
| | 60–80 | 14776712 | 185 | 0.4263 | -0.8526 | -0.8526 | 0.0238 | -0.8764 | 0.0742 | -11.8173 |
| | 80< | 5303266 | 70 | 0.4495 | -0.7997 | -0.7997 | 0.0081 | -0.8078 | 0.1199 | -6.7361 |
| TRI | 0–3 | 130505040 | 3575 | 0.9424 | -0.0594 | -0.0594 | 0.0308 | -0.0901 | 0.0205 | -4.4066 |
| | 3–6 | 150433747 | 5410 | 1.2372 | 0.2128 | 0.2128 | -0.1762 | 0.3890 | 0.0193 | 20.2064 |
| | 6–9 | 63968806 | 1400 | 0.7529 | -0.2838 | -0.2838 | 0.0502 | -0.3340 | 0.0286 | -11.6582 |
| | 9–12 | 18643271 | 305 | 0.5628 | -0.5748 | -0.5748 | 0.0229 | -0.5977 | 0.0581 | -10.2897 |
| | 12–15 | 5626230 | 67 | 0.4097 | -0.8924 | -0.8924 | 0.0090 | -0.9015 | 0.1226 | -7.3558 |
| | 15–18 | 2078584 | 35 | 0.5793 | -0.5460 | -0.5460 | 0.0024 | -0.5484 | 0.1693 | -3.2389 |
| | 18< | 1772270 | 36 | 0.6988 | -0.3584 | -0.3584 | 0.0014 | -0.3599 | 0.1669 | -2.1555 |
| Landform | Level Plain | 145490188 | 3742 | 0.8860 | -0.1210 | -0.1210 | 0.0703 | -0.1914 | 0.0202 | -9.4692 |
| | Sloping Hills | 171206942 | 5137 | 1.0336 | 0.0331 | 0.0331 | -0.0289 | 0.0620 | 0.0192 | 3.2226 |
| | Sloping Mountains | 42325701 | 1302 | 1.0597 | 0.0580 | 0.0580 | -0.0077 | 0.0657 | 0.0295 | 2.2221 |
| | Steep Plain | 4255144 | 78 | 0.6315 | -0.4597 | -0.4597 | 0.0042 | -0.4639 | 0.1136 | -4.0826 |
| | Terraced Hills | 3485600 | 268 | 2.6487 | 0.9741 | 0.9741 | -0.0157 | 0.9897 | 0.0619 | 16.0005 |
| | Terraced Mountains | 1112762 | 17 | 0.5263 | -0.6419 | -0.6419 | 0.0014 | -0.6433 | 0.2427 | -2.6504 |
| | Water | 5137555 | 284 | 1.9043 | 0.6441 | 0.6441 | -0.0127 | 0.6568 | 0.0601 | 10.9227 |
| lithology | Granite | 29450948 | 309 | 0.3614 | -1.0177 | -1.0177 | 0.0533 | -1.0710 | 0.0577 | -18.5550 |
| | Pelitic rock | 268945330 | 7850 | 1.0055 | 0.0055 | 0.0055 | -0.0143 | 0.0198 | 0.0215 | 0.9201 |
| | Siltstone | 4639738 | 87 | 0.6460 | -0.4370 | -0.4370 | 0.0044 | -0.4415 | 0.1076 | -4.1012 |
| | Marl and mixtures | 54702407 | 2088 | 1.3149 | 0.2738 | 0.2738 | -0.0556 | 0.3294 | 0.0244 | 13.5235 |
| | Carbonate rocks | 5287591 | 73 | 0.4756 | -0.7432 | -0.7432 | 0.0075 | -0.7507 | 0.1174 | -6.3922 |
| | Aeolian rock | 3270071 | 86 | 0.9060 | -0.0987 | -0.0987 | 0.0008 | -0.0996 | 0.1083 | -0.9197 |
| | Water | 5137555 | 284 | 1.9043 | 0.6441 | 0.6441 | -0.0127 | 0.6568 | 0.0601 | 10.9227 |
| | Urban areas | 1580252 | 51 | 1.1118 | 0.1060 | 0.1060 | -0.0005 | 0.1064 | 0.1404 | 0.7583 |
| Distance to faults | 0–1000 | 27392044 | 651 | 0.9909 | -0.0092 | -0.0092 | 0.0022 | -0.0113 | 0.0435 | -0.2600 |
| | 1000–2000 | 25353300 | 525 | 0.8634 | -0.1469 | -0.1469 | 0.0291 | -0.1760 | 0.0474 | -3.7105 |
| | 2000–3000 | 23982762 | 581 | 1.0100 | 0.0100 | 0.0100 | -0.0020 | 0.0120 | 0.0455 | 0.2641 |
| | 3000–4000 | 23202177 | 544 | 0.9775 | -0.0227 | -0.0227 | 0.0043 | -0.0271 | 0.0467 | -0.5789 |
| | 4000–5000 | 22042795 | 541 | 1.0233 | 0.0230 | 0.0230 | -0.0043 | 0.0273 | 0.0469 | 0.5820 |
| | 5000–6000 | 20909140 | 585 | 1.1665 | 0.1540 | 0.1540 | -0.0290 | 0.1830 | 0.0454 | 4.0299 |
| | 6000< | 230145730 | 7401 | 1.3408 | 0.2932 | 0.2932 | 0.6412 | -0.3479 | 0.0108 | -32.2323 |
| STI | 0–10 | 122777542 | 2819 | 0.7686 | -0.2632 | -0.2632 | 0.1303 | -0.3934 | 0.0223 | -17.6189 |
| | 10–20 | 96738554 | 3535 | 1.2233 | 0.2015 | 0.2015 | -0.0985 | 0.3000 | 0.0211 | 14.2490 |
| | 20–30 | 67603693 | 2291 | 1.1344 | 0.1261 | 0.1261 | -0.0357 | 0.1618 | 0.0239 | 6.7762 |
| | 30–40 | 39836701 | 1122 | 0.9428 | -0.0589 | -0.0589 | 0.0079 | -0.0668 | 0.0317 | -2.1037 |
| | 40< | 45855695 | 1061 | 0.7746 | -0.2555 | -0.2555 | 0.0361 | -0.2916 | 0.0325 | -8.9670 |
| SPI | 0–100 | 267684692 | 7415 | 0.9526 | -0.0485 | -0.0485 | 0.1184 | -0.1669 | 0.0209 | -7.9771 |
| | 100–400 | 80912199 | 2726 | 1.1587 | 0.1473 | 0.1473 | -0.0457 | 0.1929 | 0.0222 | 8.6971 |
| | 400–800 | 13764210 | 417 | 1.0419 | 0.0410 | 0.0410 | -0.0016 | 0.0427 | 0.0500 | 0.8543 |
| | 800–1200 | 4215757 | 110 | 0.8973 | -0.1083 | -0.1083 | 0.0012 | -0.1095 | 0.0958 | -1.1426 |
| | 1200–1600 | 1988524 | 49 | 0.8474 | -0.1655 | -0.1655 | 0.0008 | -0.1664 | 0.1432 | -1.1619 |
| | 1600< | 4246803 | 111 | 0.8989 | -0.1066 | -0.1066 | 0.0012 | -0.1078 | 0.0954 | -1.1297 |
| TWI | 0–4 | 6169173 | 88 | 0.4904 | -0.7125 | -0.7125 | 0.0086 | -0.7212 | 0.1070 | -6.7372 |
| | 4–6 | 211639339 | 5249 | 0.8527 | -0.1594 | -0.1594 | 0.1807 | -0.3401 | 0.0193 | -17.6145 |
| | 6–8 | 127237021 | 4504 | 1.2170 | 0.1964 | 0.1964 | -0.1212 | 0.3176 | 0.0196 | 16.2374 |
| | 8–10 | 24048685 | 895 | 1.2795 | 0.2464 | 0.2464 | -0.0197 | 0.2661 | 0.0349 | 7.6218 |
| | 10< | 3717967 | 92 | 0.8507 | -0.1617 | -0.1617 | 0.0015 | -0.1632 | 0.1047 | -1.5587 |
| Distance to stream | 0–200 | 23398203 | 936 | 1.1362 | 0.1277 | 0.1277 | -0.0152 | 0.1429 | 0.0347 | 4.1170 |
| | 200–600 | 41611589 | 1952 | 1.3324 | 0.2870 | 0.2870 | -0.0742 | 0.3611 | 0.0259 | 13.9485 |
| | 600–1200 | 55650292 | 2142 | 1.0933 | 0.0892 | 0.0892 | -0.0293 | 0.1185 | 0.0251 | 4.7224 |
| | 1200–1800 | 48654752 | 1522 | 0.8885 | -0.1182 | -0.1182 | 0.0287 | -0.1469 | 0.0284 | -5.1765 |
| | 1800–2300 | 35329567 | 945 | 0.7597 | -0.2748 | -0.2748 | 0.0416 | -0.3164 | 0.0346 | -9.1535 |
| | 2300–2800 | 30510005 | 782 | 0.7280 | -0.3174 | -0.3174 | 0.0397 | -0.3572 | 0.0376 | -9.5048 |
| | 2800< | 137873540 | 2549 | 0.5251 | -0.6441 | -0.6441 | 0.5146 | -1.1587 | 0.0238 | -48.6694 |
| NDVI | -1–0 | 6630003 | 137 | 0.7080 | -0.3453 | -0.3453 | 0.0053 | -0.3506 | 0.0860 | -4.0779 |
| | 0–0.2 | 7689845 | 287 | 1.2788 | 0.2459 | 0.2459 | -0.0059 | 0.2518 | 0.0598 | 4.2090 |
| | 0.2–0.4 | 18076725 | 798 | 1.5125 | 0.4138 | 0.4138 | -0.0267 | 0.4405 | 0.0368 | 11.9741 |
| | 0.4–0.6 | 81792466 | 3387 | 1.4188 | 0.3498 | 0.3498 | -0.1266 | 0.4764 | 0.0207 | 22.9681 |
| | 0.6–0.8 | 255713304 | 6187 | 0.8290 | -0.1875 | -0.1875 | 0.3242 | -0.5118 | 0.0195 | -26.3010 |
| | 0.8< | 3116202 | 32 | 0.3518 | -1.0446 | -1.0446 | 0.0055 | -1.0501 | 0.1770 | -5.9312 |
| Land use | Water | 10128289 | 326 | 1.1089 | 0.1034 | 0.1034 | -0.0030 | 0.1065 | 0.0562 | 1.8928 |
| | Trees | 257597190 | 5378 | 0.7193 | -0.3295 | -0.3295 | 0.4871 | -0.8166 | 0.0192 | -42.4706 |
| | Flooded vegetation | 1545249 | 79 | 1.7614 | 0.5661 | 0.5661 | -0.0032 | 0.5693 | 0.1129 | 5.0412 |
| | Crops | 32228327 | 1027 | 1.0979 | 0.0934 | 0.0934 | -0.0093 | 0.1027 | 0.0328 | 3.1310 |
| | Built area | 7314103 | 296 | 1.3943 | 0.3324 | 0.3324 | -0.0079 | 0.3403 | 0.0589 | 5.7743 |
| | Bare ground | 63795611 | 3709 | 2.0031 | 0.6947 | 0.6947 | -0.2320 | 0.9267 | 0.0203 | 45.7530 |
| | Clouds | 148744 | 6 | 1.3898 | 0.3291 | 0.3291 | -0.0002 | 0.3293 | 0.4084 | 0.8063 |
| | Rangeland | 212439 | 7 | 1.1352 | 0.1269 | 0.1269 | -0.0001 | 0.1269 | 0.3781 | 0.3357 |
| Distance to rod | 0–200 | 81857744 | 3938 | 1.5211 | 0.4194 | 0.4194 | -0.1936 | 0.6130 | 0.0203 | 30.1899 |
| | 200–600 | 82157525 | 2554 | 0.9829 | -0.0173 | -0.0173 | 0.0058 | -0.0230 | 0.0228 | -1.0092 |
| | 600–1200 | 70851246 | 1846 | 0.8238 | -0.1938 | -0.1938 | 0.0481 | -0.2419 | 0.0257 | -9.4127 |
| | 1200–1800 | 44619411 | 1038 | 0.7355 | -0.3071 | -0.3071 | 0.0413 | -0.3485 | 0.0327 | -10.6440 |
| | 1800–2300 | 25977658 | 551 | 0.6706 | -0.3995 | -0.3995 | 0.0283 | -0.4278 | 0.0438 | -9.7687 |
| | 2300–2800 | 18746343 | 327 | 0.5515 | -0.5951 | -0.5951 | 0.0272 | -0.6222 | 0.0562 | -11.0708 |
| | 2800< | 48818021 | 574 | 0.3718 | -0.9895 | -0.9895 | 0.1056 | -1.0951 | 0.0430 | -25.4915 |
| Peak rainfall Intensity | 13–15 | 23809786 | 244 | 0.3755 | -0.9796 | -0.9796 | 0.0495 | -1.0291 | 0.0649 | -15.8467 |
| | 15–17 | 102293235 | 2750 | 0.9849 | -0.0152 | -0.0152 | 0.0072 | -0.0223 | 0.0231 | -0.9678 |
| | 17–19 | 126644544 | 3960 | 1.1456 | 0.1359 | 0.1359 | -0.1019 | 0.2378 | 0.0216 | 11.0193 |
| | 19–22 | 64270925 | 1699 | 0.9685 | -0.0320 | -0.0320 | 0.0080 | -0.0400 | 0.0271 | -1.4778 |
| | 22< | 56009382 | 2175 | 1.4227 | 0.3526 | 0.3526 | -0.0951 | 0.4477 | 0.0248 | 18.0634 |
| Average rainfall Intensity | 2.6–3.2 | 30873234 | 487 | 0.5571 | -0.5850 | -0.5850 | 0.0392 | -0.6242 | 0.0464 | -13.4533 |
| | 3.2–3.4 | 43648880 | 1067 | 0.8633 | -0.1470 | -0.1470 | 0.0179 | -0.1649 | 0.0323 | -5.1074 |
| | 3.4–3.6 | 68230659 | 2361 | 1.2221 | 0.2005 | 0.2005 | -0.0509 | 0.2515 | 0.0234 | 10.7690 |
| | 2.6–3.8 | 230545414 | 6655 | 1.0195 | 0.0193 | 0.0193 | -0.0319 | 0.0512 | 0.0201 | 2.5429 |
| | 3.8< | 142482458 | 4173 | 1.0344 | 0.0338 | 0.0338 | -0.0214 | 0.0552 | 0.0199 | 2.7744 |